\documentclass[11pt,a4paper]{article}

\usepackage[dvipsnames]{xcolor}
\definecolor{iccvblue}{rgb}{0.196,0.553,0.87}
\usepackage{graphicx}
\usepackage{booktabs}
\usepackage[numbers,sort&compress]{natbib}
\RequirePackage{etoolbox}

\usepackage[
pdftex,
pdfauthor={Giacomo Meanti, Thomas Ryckeboer, Michael Arbel, Julien Mairal},
pdftitle={Unsupervised Imaging Inverse Problems with Diffusion Distribution Matching},
pagebackref,breaklinks,colorlinks,allcolors=iccvblue
]{hyperref}
\usepackage[utf8]{inputenc} % allow utf-8 input
\usepackage{url}            % simple URL typesetting
\usepackage[left=2.5cm, right=2cm, bottom=2cm, top=2cm]{geometry}
\usepackage{adjustbox}
\usepackage{amsfonts}       % blackboard math symbols
\usepackage{amsthm}
\usepackage{amssymb}
\usepackage{mathtools}
\usepackage{float}
\usepackage{nicefrac}       % compact symbols for 1/2, etc.
\usepackage{microtype}      % microtypography
\usepackage{xspace}
\usepackage{xparse}         % for \NewDocumentCommand
\usepackage{stmaryrd}       % For \llbracket etc
\usepackage{placeins}
\usepackage{subcaption}
\usepackage[inline]{enumitem}
\usepackage{bm}
\usepackage{commath} % for \dif,  \od[2]{f}{x},  \pd[2]{f}{x}
\usepackage{braket} 	% For \Set{ \ldots  | \ldots  }
\usepackage{tabularray}  % For author affiliations
\usepackage{makecell}
\usepackage{multirow}
\usepackage[page]{appendix} % print appendices title
\usepackage{import}
\usepackage{capt-of}
\usepackage[capitalize]{cleveref}
\usepackage{siunitx}
\robustify\bfseries
\NewDocumentCommand\fmap{g}{\phi\IfNoValueF{#1}{(#1)}}
\DeclareDocumentCommand\cme{g}{μ_p\IfNoValueF{#1}{(#1)}}

\DeclareDocumentCommand\ldm{g}{\tilde{\bm{x}}\IfNoValueF{#1}{_{#1}}} % (i-th) landmark
\DeclareDocumentCommand\ldmY{g}{\tilde{\bm{y}}\IfNoValueF{#1}{_{#1}}} % (i-th) landmark

\NewDocumentCommand\kmat{O{n}}{K_{#1}}

\NewDocumentCommand\rest{O{\lambda}}{A_{#1}}	% Regularized estimator
\NewDocumentCommand\eest{O{\lambda}}{\hat{A}_{#1}}	% Empirical estimator
\NewDocumentCommand\nysest{O{\lambda}}{\hat{A}_{m,#1}^{\text{KRR}}}
\NewDocumentCommand\rrrest{O{\lambda}}{\hat{A}_{m,#1}^{\text{RRR}}}
\NewDocumentCommand\pcrest{O{\lambda}}{\hat{A}_{m}^{\text{PCR}}}
\NewDocumentCommand\fkrrest{O{\lambda}}{\hat{A}_{#1}^{\text{KRR}}}
\NewDocumentCommand\frrrest{O{\lambda}}{\hat{A}_{#1}^{\text{RRR}}}
\NewDocumentCommand\fpcrest{O{\lambda}}{\hat{A}_{}^{\text{PCR}}}

\NewDocumentCommand\krr{O{\lambda}}{A_{#1}}
\NewDocumentCommand\nkrr{O{\lambda}}{\tilde{A}_{#1}}
\NewDocumentCommand\ekrr{O{\lambda}}{\hat{A}_{#1}}
\NewDocumentCommand\rrr{O{\lambda}}{{A}_{#1}^{\text{RRR}}}
\NewDocumentCommand\pcr{O{\lambda}}{{A}_{#1}^{\text{PCR}}}
\NewDocumentCommand\npcr{O{\lambda}}{\hat{A}_{m,#1}^{\text{PCR}}}

\DeclareDocumentCommand\cme{g}{\mu_p\IfNoValueF{#1}{(#1)}}

\newcommand{\whp}{w.h.p.}

\newcommand\restr[2]{{% we make the whole thing an ordinary symbol
  \left.\kern-\nulldelimiterspace % automatically resize the bar with \right
  #1 % the function
  \vphantom{\big|} % pretend it's a little taller at normal size
  \right|_{#2} % this is the delimiter
  }}

\DeclareMathOperator*{\argmin}{arg\,min}

\DeclareMathOperator*{\trc}{tr}

\newcommand{\V}[1]{\symbf{#1}} % for pdflatex

\newcommand{\bE}{\mathbb{E}}

\let\norm\relax
\DeclarePairedDelimiter{\norm}{\lVert}{\rVert}

\DeclarePairedDelimiterX{\infdivx}[2]{(}{)}{%
	#1\;\delimsize\|\;#2%
}
\DeclareMathOperator{\kldivop}{KL}
\newcommand{\kld}{\kldivop\infdivx}

\DeclareFontFamily{U}{matha}{\hyphenchar\font45}
\DeclareFontShape{U}{matha}{m}{n}{
<-6> matha5 <6-7> matha6 <7-8> matha7
<8-9> matha8 <9-10> matha9
<10-12> matha10 <12-> matha12
}{}
\DeclareSymbolFont{matha}{U}{matha}{m}{n}

\DeclareFontFamily{U}{mathx}{\hyphenchar\font45}
\DeclareFontShape{U}{mathx}{m}{n}{
<-6> mathx5 <6-7> mathx6 <7-8> mathx7
<8-9> mathx8 <9-10> mathx9
<10-12> mathx10 <12-> mathx12
}{}
\DeclareSymbolFont{mathx}{U}{mathx}{m}{n}

\DeclareMathDelimiter{\vvvert} {0}{matha}{"7E}{mathx}{"17}%
\DeclarePairedDelimiterX{\normiii}[1]
{\vvvert}
{\vvvert}
{\ifblank{#1}{\:\cdot\:}{#1}}

\usepackage{pgffor}
\foreach \x in {a,...,z}{
  \expandafter\xdef\csname V\x \endcsname{\noexpand\ensuremath{\noexpand\V{\x}}}
}
\foreach \x in {A,...,Z}{
  \expandafter\xdef\csname V\x \endcsname{\noexpand\ensuremath{\noexpand\V{\x}}}
  \expandafter\xdef\csname c\x \endcsname{\noexpand\ensuremath{\noexpand\mathcal{\x}}}
  \expandafter\xdef\csname f\x \endcsname{\noexpand\ensuremath{\noexpand\mathfrak{\x}}}
}

\Crefname{equation}{Eq.}{Eqs.}
\crefname{equation}{eq.}{eqs.}

\renewcommand{\whp}{\omega}
\newcommand{\wopt}{\whp_{*}}
\newcommand{\what}{\hat{\whp}}
\newcommand{\fw}{\cA_{\whp}}
\newcommand{\hatfw}{\hat{\cA}_{\whp}}
\newcommand{\fwhat}{\cA_{\what}}

\newcommand{\fwopt}{\cA_{\wopt}}
\newcommand{\dY}[2][]{%
	\ifthenelse{\equal{#1}{}}%
		{\mathbb{Y}_{#2}}%
		{\mathbb{Y}^{(#1)}_{#2}}%
}
\newcommand{\dX}{\mathbb{X}}
\newcommand{\dYw}{\dY{\whp}}

\newcommand{\dYopt}{\dY{\wopt}}

\DeclareMathOperator*{\ikldivop}{IKL}
\newcommand{\IKL}{\ikldivop\infdivx}
\usepackage[most]{tcolorbox}

\definecolor{customBlue}{RGB}{18,75,126}
\colorlet{titleCol}{customBlue}	% for lemmas, proofs etc.
\colorlet{titleThmCol}{titleCol} % for theorems
\colorlet{backCol}{titleCol!08!white}
\colorlet{backThmCol}{titleThmCol!08!white}

\tcbset{
  compactdefaultstyle/.style={
	breakable,
	theorem style=plain,
	enhanced,
	sharp corners,
	frame hidden,
	colback=backCol,
	opacityfill=0, 
	fontupper=\itshape,
	fonttitle=\bfseries\upshape,
	coltitle=black, 
	boxrule=0pt,
	before skip=1mm,
	after skip=1mm,
	right = 0mm, 
	left = -1mm, 
	top = 0mm,
	bottom = 0mm,
  },
  defaultthmstyle/.style={
	breakable,
	theorem style=plain,
	enhanced,
	sharp corners,
	frame hidden,
	colback=backCol,
	opacityfill=0, 
	fontupper=\itshape,
	fonttitle=\bfseries\upshape,
	coltitle=black, 
	boxrule=0pt,
	before skip=2mm,
	after skip=2mm,
	right = 0mm, 
	left = -1mm, 
	top = 1mm,
	bottom = 1mm,
  },
  defaultproofstyle/.style={
	breakable,
	theorem style=plain,
	terminator sign={.},  % End of title, before the content of the box (to separate because it's inlined)
	enhanced,
	sharp corners,
	frame hidden,
	colback=backCol,
	opacityfill=0, 
	fontupper=\upshape,
	fonttitle=\itshape,
	separator sign={:\ }, 
	coltitle=black, 
	boxrule=0pt,
	before skip=2mm,
	after skip=2mm,
	right = 0mm, 
	left = -1mm, 
	top = 1mm,
	bottom = 1mm,
  },
  thmstyle/.style={
	breakable,
	sharp corners,
	colbacktitle=titleThmCol,
	colback=backThmCol,
	colframe=titleThmCol, 
	fonttitle=\upshape\bfseries\hypersetup{linkcolor=white,citecolor=white},
	left = 2mm, 
	right = 2mm, 
	before skip=2mm,
	after skip=2mm,
  },
  lemmastyle/.style={
	breakable,
	enhanced,
	theorem style=plain,
	frame hidden,
	sharp corners,
	colback=backCol,
	boxrule=0pt,
	before skip=2mm,
	after skip=2mm,
	right = 2mm, 
	borderline={0.4mm}{0pt}{titleCol},
	fonttitle=\bfseries,
	coltitle=titleCol,
  },
  defstyle/.style={
	breakable,
	theorem style=plain,
	enhanced,
	sharp corners,
	frame hidden,
	colback=backCol,
	coltitle=titleCol,
	boxrule=0pt,
	borderline west={0.8mm}{0pt}{backCol},
	fonttitle=\upshape\bfseries\hypersetup{citecolor=titleCol},
	fontupper=\upshape,
	before skip=2mm,
	after skip=2mm,
	left = 3mm, 
  },
  proofstyle/.style={
	breakable,
	enhanced,
	theorem style=plain,
	frame hidden,
	sharp corners,
	colback=backCol,
	boxrule=0pt,
	before skip=2mm,
	after skip=2mm,
	right = 2mm, 
	borderline west={0.4mm}{0pt}{titleCol},
	fonttitle=\bfseries,
	coltitle=titleCol,
  }, 
  bluewestlinestyle/.style={
	breakable,
	theorem style=plain,
	enhanced,
	sharp corners,
	frame hidden,
	coltitle=customBlue,
	boxrule=0pt,
	borderline west={0.8mm}{0pt}{customBlue},
	fonttitle=\upshape\bfseries\hypersetup{citecolor=customBlue,linkcolor=customBlue},
	before skip=2mm,
	after skip=2mm,
	left = 3mm, 
	opacityfill=0, 
	bottom=0mm, 
	top=0mm, 
	toptitle=0mm,
	bottomtitle=0mm,
	right = 0mm, 
  },
  redvioletwestlinestyle/.style={
	breakable,
	theorem style=plain,
	enhanced,
	sharp corners,
	frame hidden,
	colback=black!03!white,
	coltitle=RedViolet,
	boxrule=0pt,
	borderline west={0.8mm}{0pt}{RedViolet},
	fonttitle=\upshape\bfseries\hypersetup{citecolor=RedViolet,linkcolor=RedViolet},
	before skip=2mm,
	after skip=2mm,
	left = 3mm, 
	opacityfill=0, 
	bottom=0mm, 
	top=0mm, 
	toptitle=0mm,
	bottomtitle=0mm,
	right = 0mm, 
  },
  greenwestlinestyle/.style={
	breakable,
	theorem style=plain,
	enhanced,
	sharp corners,
	frame hidden,
	coltitle=ForestGreen,
	boxrule=0pt,
	borderline west={0.8mm}{0pt}{ForestGreen},
	fonttitle=\upshape\bfseries\hypersetup{citecolor=ForestGreen,linkcolor=ForestGreen},
	before skip=2mm,
	after skip=2mm,
	left = 3mm, 
	opacityfill=0, 
	bottom=0mm, 
	top=0mm, 
	toptitle=0mm,
	bottomtitle=0mm,
	right = 0mm, 
  },
}

\tcbset{todobox/.style={
  title={Some work is needed here}, 
  colframe=red!70!black,
  colback=red!10,
  fonttitle=\bfseries, 
  coltext=red!70!black, 
  breakable,
  boxed title style={
    outer arc=0pt,
    arc=0pt,
    top=0pt,
    bottom=0pt,
    },
  }
}

\tcbset{warnbox/.style={
  colframe=black,
  colback=yellow!20,
  breakable,
  boxed title style={
    outer arc=0pt,
    arc=0pt,
    top=0pt,
    bottom=0pt,
    },
  }
}
\tcbset{infobox/.style={
  colframe=blue,
  colback=blue!05,
  breakable,
  boxed title style={
    outer arc=0pt,
    arc=0pt,
    top=0pt,
    bottom=0pt,
    },
  }
}

\makeatletter
\def\@LN@depthbox{%
  \ifdim\@tempdima = -1000pt
  \else
    \dp\@tempboxa=\@tempdima
    \nointerlineskip \kern-\@tempdima 
  \fi
  \box\@tempboxa
  } 
\makeatother

\newtcbtheorem[crefname={theorem}{theorems},Crefname={Theorem}{Theorems},number within=section]{ttheorem}{Theorem}{defaultthmstyle}{r}
\newtcbtheorem[crefname={lemma}{lemmas},Crefname={Lemma}{Lemmas},use counter from=ttheorem]{tlemma}{Lemma}{defaultthmstyle}{r}
\newtcbtheorem[crefname={proposition}{propositions},Crefname={Proposition}{Propositions},use counter from=ttheorem]{tprop}{Proposition}{defaultthmstyle}{r}
\newtcbtheorem[crefname={proof}{proofs},Crefname={Proof}{Proofs},number within=section]{tproof}{Proof}{defaultproofstyle}{pf}
\newtcbtheorem[crefname={remark}{remarks},Crefname={Remark}{Remarks},number within=section, ]{tremark}{Remark}{defaultthmstyle}{a}
\newtcbtheorem[crefname={assumption}{assumptions},Crefname={Assumption}{Assumptions},use counter from=ttheorem, ]{tassumption}{Assumption}{compactdefaultstyle}{a}

\newenvironment{tproofof*}[2]{
\begin{tproof*}[title={Proof of \Cref{#1}:}]{}{#2}
}{
\null\hfill$\square$
\end{tproof*}
}

\graphicspath{{./figures/}}

\makeatletter
\patchcmd{\@setref}{\bfseries ??}{\bfseries\color{red} undefined Label}{}{}
\makeatother
\makeatletter
\patchcmd{\@@setcref}         {??}{\color{red} undefined Label}{}{}
\patchcmd{\@@setcref}         {??}{\color{red} undefined Label}{}{}
\patchcmd{\@@setcrefrange}    {??}{\color{red} undefined Label}{}{}
\patchcmd{\@@setcrefrange}    {??}{\color{red} undefined Label}{}{}
\patchcmd{\@@setcrefrange}    {??}{\color{red} undefined Label}{}{}
\patchcmd{\@@setcrefrange}    {??}{\color{red} undefined Label}{}{}
\patchcmd{\@@setcrefrange}    {??}{\color{red} undefined Label}{}{}
\patchcmd{\@@setcrefrange}    {??}{\color{red} undefined Label}{}{}
\patchcmd{\@@setnamecref}     {??}{\color{red} undefined Label}{}{}
\patchcmd{\@@setnamecref}     {??}{\color{red} undefined Label}{}{}
\patchcmd{\@@setcpageref}     {??}{\color{red} undefined Label}{}{}
\patchcmd{\@@setcpageref}     {??}{\color{red} undefined Label}{}{}
\patchcmd{\@@setcpagerefrange}{??}{\color{red} undefined Label}{}{}
\patchcmd{\@@setcpagerefrange}{??}{\color{red} undefined Label}{}{}
\patchcmd{\@@setcpagerefrange}{??}{\color{red} undefined Label}{}{}
\patchcmd{\@@setcpagerefrange}{??}{\color{red} undefined Label}{}{}
\patchcmd{\@@setcpagerefrange}{??}{\color{red} undefined Label}{}{}
\patchcmd{\@@cref}            {??}{\color{red} undefined Label}{}{}
\makeatother

\setlist*[enumerate]{label=\roman*)}

\title{Unsupervised Imaging Inverse Problems \\ with Diffusion Distribution Matching}
\author{{\large Giacomo Meanti~~~Thomas Ryckeboer~~~Michael Arbel~~~Julien Mairal} \\
{\normalsize Univ. Grenoble Alpes, Inria, CNRS, Grenoble INP, LJK}, 
{\normalsize 38000 Grenoble, France} \\
{\normalsize \tt firstname.lastname@inria.fr }
}
\date{}

\begin{document}
	\maketitle
    \begin{abstract}

This work addresses image restoration tasks through the lens of inverse problems using unpaired datasets. In contrast to traditional approaches---which typically assume full knowledge of the forward model or access to paired degraded and ground-truth images---the proposed method operates under minimal assumptions and relies only on small, unpaired datasets. This makes it particularly well-suited for real-world scenarios, where the forward model is often unknown or mis-specified, and collecting paired data is costly or infeasible.
The method leverages conditional flow matching to model the distribution of degraded observations, while simultaneously learning the forward model via a distribution-matching loss that arises naturally from the framework. Empirically, it outperforms both single-image blind and unsupervised approaches on deblurring and non-uniform point spread function (PSF) calibration tasks. It also matches state-of-the-art performance on blind super-resolution.
%To further demonstrate its practical relevance, the method is applied to lens calibration—a real-world task that traditionally requires extensive experiments and specialized hardware. In contrast, our approach achieves comparable results with minimal data collection, highlighting its efficiency and broad applicability.
%This work addresses image restoration tasks, seen through the framework of inverse problems, with unpaired datasets.
%Unlike more standard setups for solving inverse problems which require full knowledge of the forward model or a dataset of ground truth and measurement pairs, the proposed method requires minimal assumptions on the forward model and small, unpaired datasets.
%This approach is in fact particularly useful for real-world restoration tasks where the forward model cannot be known precisely and paired-data collection is expensive even when it is at all possible.
%We use conditional flow matching to model the degraded data distribution and explicitly learn the forward model with a naturally arising distribution matching loss.
%The proposed method outperforms both single-image blind and unsupervised approaches on deblurring and non-uniform psf calibration as well as performs on par with SoTA on blind super-resolution.
We also showcase the effectiveness of our method with a proof of concept for lens calibration: a real-world application traditionally requiring time-consuming experiments and specialized equipment. In contrast, our approach achieves this with minimal data acquisition effort. Code available: \url{https://github.com/inria-thoth/ddm4ip}.

% v1. 08/05
% Image reconstruction tasks from deblurring and calibration to low-light enhancement can often be framed as inverse problems.
% Solving them requires either full knowledge of the forward model, or a dataset of ground truth and measurement pairs.
% The former requirement is often unfeasible and leads to suboptimal reconstruction due to mis-specification. The latter is expensive even when it is at all possible.
% Without paired supervision or an explicit forward model, solving inverse problems becomes significantly more challenging. Traditionally this has been addressed with strong priors to regularize the problem, enabling reconstruction from individual examples.
% In this work we demonstrate how to accurately estimate the true degradation operator with minimal assumptions on the forward model and by leveraging small (down to single images), unpaired datasets.
% We use conditional flow matching to model the degraded data distribution and explicitly learn the forward model with a naturally arising distribution matching loss. 
% The proposed method outperforms both single-image blind and unsupervised approaches on uniform and non-uniform deblurring as well as super-resolution tasks, narrowing the gap to non-blind algorithms.
% We also showcase the effectiveness of our method with a proof of concept for lens calibration: a real-world application traditionally requiring time-consuming experiments and specialized equipment. In contrast, our approach achieves this with minimal data acquisition effort.
\end{abstract}

    \begin{figure}[ht!]
    	\centering
    	%	\vspace*{-0.5cm}
    	\includegraphics[width=0.93\textwidth]{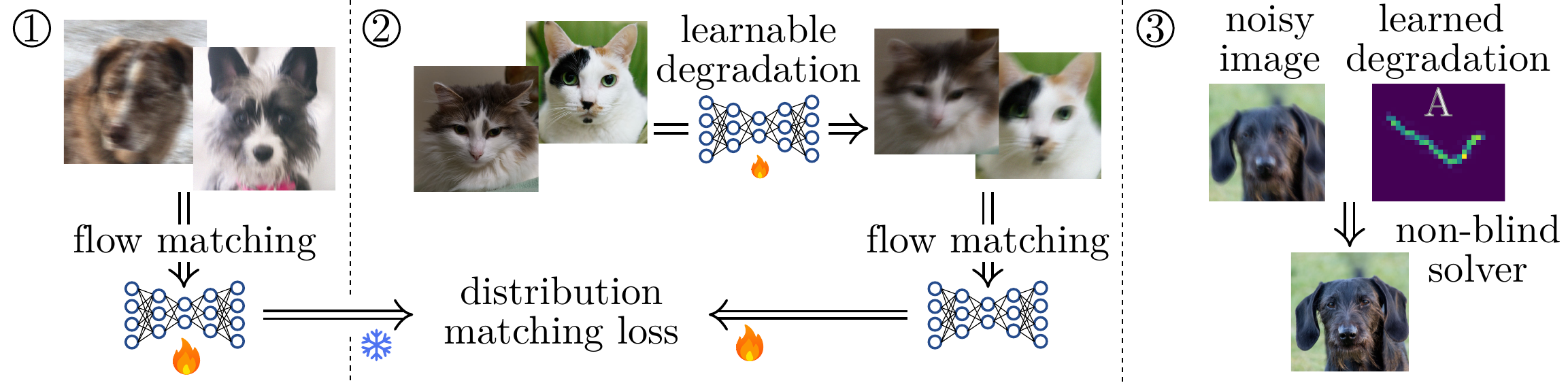}
    	\caption{Learning degradation operators by matching distributions.}
    	%	\vspace*{0.5cm}
    	\label{fig:teaser}
    \end{figure}
    
    \section{Introduction}\label{sec:intro}

%Inverse problems arise across various domains where a noisy, lossy process corrupts data that we aim to recover. More formally, consider the forward model: 
%\begin{equation}\label{eq:invp}%
%	y = \cA(x) + \epsilon, \quad \epsilon \sim \cN(0, \sigma^2 I).
%\end{equation}
%The goal is to reconstruct $x$ from measurements $y$, a challenging task when the corruption operator $\cA$ is lossy, i.e.~not easily invertible. 
%In practice, the forward operator $\cA$ can represent very diverse processes both linear and non-linear, including blur operators, inpainting masks, the Radon transform, and many more.

Inverse problems arise in many domains where observed data is the result of a noisy and potentially lossy transformation of an underlying signal we wish to recover. More formally, the degradation process is modeled as follows:
\begin{equation}\label{eq:invp}
	y = \cA(x) + \epsilon, \quad \epsilon \sim \cN(0, \sigma^2 I),
\end{equation}
where $\cA$ is the forward operator and $\epsilon$ denotes additive Gaussian noise. The objective is to recover the original signal $x$ from the measurements $y$. In practice, $\cA$ may represent a wide range of linear or nonlinear transformations, such as blur kernels, inpainting masks, the Radon transform, and many others.

% Categorization:
% - cA is known
%  + PnP
%  + clean data is available, cA used to generate a paired dataset for learning with supervised techniques
% - cA is unknown
%  + cA is fixed
%   - our method (learn cA from the data)
%  + cA is data-dependent
%   - make assumptions on the distribution of possible cA to generate a paired dataset
%   - make assumptions on the distribution of possible cA to learn a prior over cA
%   - our method (learn cA from the data)

To introduce the setting of the current work, we categorize the reconstruction methods based on 
\begin{enumerate*}
	\item the type and availability of data,
	\item the extent of prior knowledge about~$\cA$.
\end{enumerate*}
When $\cA$ is fully known, two main classes of reconstruction approaches are typically used. The first is plug-and-play (PnP) methods, which apply iterative algorithms to recover signal $x$ from a single measurement $y$ by maximizing the posterior distribution $p(x \mid y, \cA) \propto p(y \mid x, \cA) p(x)$. These methods alternate between a data-fidelity term (dependent on $\cA$) and a prior term. The second class leverages supervised learning: if a clean dataset is available, it can be paired with synthetic measurements generated via $\cA$ to train a reconstruction model in a supervised fashion.

%When $\cA$ is fully available there are two main options for reconstruction: plug-and-play (PnP) methods consist of an iterative algorithm which restores single data-points $y$ by maximizing posterior probability $p(x \mid y, \cA) \propto p(y\mid x, \cA) p(x)$, iterating between a data-fidelity term (which depends on $\cA$) and a prior term; if instead a clean dataset is available it can be used in conjunction with $\cA$ to generate a paired dataset, followed by supervised learning techniques to learn a reconstruction model.
However, complete knowledge of the forward operator $\cA$ is rarely available in practice. For instance, in super-resolution it is common to assume bicubic downsampling to generate low-resolution images from high-resolution ones, yet models trained under this assumption often fail on real-world images where the actual downsampling kernel differs~\cite{gu2019blind, zhang23sr}.

%However, full knowledge of the forward operator is rarely available in practice. For example in super-resolution, the assumption of a bicubic downsampling filter is common, yet models trained on such data fail in real-world scenarios where different downsampling filters occur~\cite{gu2019blind, zhang23sr}.
When $\cA$ is unknown or only partially specified, the problem becomes more challenging and is called \emph{blind}. One strategy is to collect (or synthetically generate) a dataset that includes corrupted samples produced from a range of possible $\cA$ instances. A supervised model is then trained to generalize across these degradations. An alternative approach is to jointly infer both the clean signal and the unknown operator by maximizing the joint posterior $p(x, \cA \mid y) \propto p(y \mid x) p(x) p(\cA)$. This formulation requires priors on both the data distribution $p(x)$ and the corruption process $p(\cA)$.
%When $\cA$ is unknown or under-specified, the problem is harder and is referred to as \emph{blind}.
%A first approach is to collect (or synthetically generate) a dataset which includes corrupted samples generated from all possible desired $\cA$, and then used supervised learning techniques to learn a generic model for reconstruction.
%An alternative is to maximize posterior probability $p(x, \cA \mid y) \propto p(y\mid x) p(x) p(\cA)$ such that for any corrupted sample, both the corresponding operator $\cA$ and the clean image are learned. This latter approach requires both the data prior $p(x)$ and a prior on the corruptions $p(\cA)$.
Both approaches implicitly rely on some prior knowledge or assumptions about $\cA$, either through heuristic degradation models or learned probabilistic priors. Consequently, when the true degradation deviates from these assumptions---which is often the case in real-world scenarios---the reconstruction performance degrades significantly. Although these methods typically require very little input data at test time (sometimes even a single image), their robustness to real-world variations is far from guaranteed.
%Clearly both approaches require some prior knowledge about the relevant operators $\cA$, either through degradation heuristics or through a probabilistic prior. Therefore, whenever the true forward operator deviates from the (often narrow) prior used for training, the learned model is likely to fail.
%Despite requiring very few data points at test time—often making them \emph{single-image} methods—robustness is far from guaranteed.

The natural alternative, which we adopt in this work, is to learn information about $\cA$ directly from data: specifically, from one dataset of corrupted images and a separate dataset of clean images. Crucially, these datasets do not need to contain corresponding clean-corrupted pairs, which significantly simplifies data collection. During training, the unpaired clean images serve as a reference for what the restored outputs should resemble—effectively providing a model for $p(x)$.
%The natural alternative, which we adopt in this work, is to learn information about $\cA$ directly from a dataset of corrupted images on the one hand, and a dataset of clean images on the other hand. Such a dataset does not need to have corresponding pairs of clean images, thus making it much easier to obtain from a data collection standpoint. Unpaired clean images are also used in the training process to provide a reference for what the restored data should look like (essentially providing $p(x)$). 
This unpaired training setup is often referred to as \emph{unsupervised}, and has been successfully applied to in-the-wild image restoration tasks~\cite{ntire20,wolf21deflow,sim20otcyclegan}, where the degradation process is unknown and may involve multiple, complex transformations. 
%This unpaired procedure is sometimes known as \emph{unsupervised} and has been applied to in-the-wild image restoration tasks~\cite{ntire20,wolf21deflow,sim20otcyclegan} where the corrupted data distribution is characterized by multiple unknown degradations. 
Most existing algorithms operating in this unpaired regime use an implicit model of the degradation process: a neural network is trained to map corrupted inputs to clean outputs, without explicitly modeling the underlying forward operator.

The method we propose learns the correct degradation %starting from unpaired clean and noisy datasets. It does so efficiently 
by learning an explicit operator $\cA$ and needs a relatively small (hundreds to thousands) number of training points. By using diffusion models and an efficient distribution matching algorithm it outperforms both single-image methods and other unsupervised ones. It comprises two distinct steps: the first one is to learn a representation of the noisy distribution, the second is to learn the best degradation operator $\cA$ such that the clean samples corrupted by $\cA$ match \emph{in distribution} the noisy data prior. Finally the learned corruption can be used in a third step for non-blind restoration.  
%The method we propose uses similar concepts to the unsupervised inverse problems literature: 
Our contributions are threefold: we begin by devising a principled method for solving imaging inverse problems which only relies on unpaired image data from clean and corrupted distributions, without knowledge of the degradation operator. We prove that under a non-degeneracy assumption on the clean distribution, the true operator can be identified. We then show how this method can provide precise estimates of the degradation operator, without making any distributional assumptions on the operator itself. In particular we focus on the tasks of uniform and non-uniform deblurring. 
Finally, since the estimates we obtain are considerably more precise than those of single-image methods, we show how our approach can be used as part of a pipeline for camera lens calibration -- where accuracy is essential.

The rest of the paper is organized as follows. In \cref{sec:bg}, we provide some necessary background on inverse problems with unknown degradations and on our algorithm. In \cref{sec:method}, we detail the different steps of our algorithm, whereas the last section is devoted to experiments of increasing complexity on estimating blur kernels.

\begin{table}
	\centering
	\caption{Summary of tasks for solving inverse problems without knowing the forward operator. The headings refer to what is known about the data and the forward operator. Regarding the latter, we note when the method works on data corrupted by a single (unknown) operator, operators from a predefined distribution, or by a collection of unknown operators. The references are not meant to be exhaustive but just to provide examples.}
	\label{tbl:setting}
	\begin{tabular}{llll}
		\toprule
		Data & $\cA$ & Method & References \\
		\midrule
		prior on $\cX$ & single known & PnP & \cite{pnp, romano17red, zhang2021plug, zhu23diffpir} \\
		paired & single unknown & supervised & \cite{wang18esrgan} \\
		prior on $\cX$ & prior on $\cA$ & blind PnP & \cite{chung2023parallel,yash24kerneldiff,laroche2024fast,yang2024dynamic} \\
		paired & prior on $\cA$ & supervised & \cite{wang21realesrgan,zhang21bsrgan,gu2019blind} \\
		unpaired & multiple unknown & unsupervised & \cite{wolf21deflow,sim20otcyclegan,luo22pdmsr} \\
		unpaired & functional form & \multicolumn{2}{c}{\emph{this paper}} \\
		\bottomrule
	\end{tabular}
\end{table}
%
%\begin{table}
%	\centering
%	\begin{tabular}{llll}
%		\toprule
%		Setup & $\cA$ data dependent? & Dataset & References \\
%		\midrule
%		Supervised & no & paired & \cite{wang18esrgan} \\
%		Supervised & yes & paired & \cite{wang21realesrgan,zhang21bsrgan,gu2019blind} \\
%		Blind      & yes & $\cY$ + priors & \cite{chung2023parallel,yash24kerneldiff,laroche2024fast,yang2024dynamic} \\
%		Unsupervised & yes & unpaired $\cX, \cY$ & \cite{wolf21deflow,sim20otcyclegan,luo22pdmsr} \\
%		This paper & no & unpaired & \cite{sim20otcyclegan} \\
%		\bottomrule
%	\end{tabular}
%\end{table}

%The alternative we propose to the single image paradigm is made of two parts. As input we assume we have access to a clean and a noisy dataset. The clean one acts as reference for what the restored images should look like, while the noisy one allows to learn the corruption distribution. To learn a corruption operator $\hat{\cA}$ then we can attempt to minimize the distance between the available corrupted dataset and the clean one corrupted by $\hat{\cA}$. When their distance is zero, then $\hat{\cA}$ will be equivalent to the true corruption. If paired data is available this can be done with standard supervised losses~\cite{sun2015learning, fang2023self}, the main difficulty for us is to accurately minimize the distance between distributions as available from small datasets. 
    \section{Background}\label{sec:bg}

%We propose an algorithm to solve imaging inverse problems by learning the unknown forward operator. This is achieved by adapting the distribution of a clean dataset to match the reference distribution of a noisy dataset. 
%The algorithm we use for distribution matching is very similar to the one used in DreamFusion~\cite{poole22dreamfusion} and DiffInstruct~\cite{luo23diffinstruct} in very different settings. 
%We briefly review the literature for blind and unsupervised inverse problems in imaging, as well as of the different forms of deblurring which we will be tackling.
We briefly review the literature on blind and unsupervised inverse problems in imaging, along with the various forms of deblurring addressed in this work.

\subsection{Unsupervised inverse problems}

Solving inverse problems without access to the forward operator is inherently challenging and can be approached from multiple perspectives. In \cref{tbl:setting} we provide a framework to clarify the various settings considered in the literature. Unlike many of the methods discussed below, our approach assumes a fixed degradation operator $\cA$ which does not vary across measurements $y$. While this restricts generality, it enables higher reconstruction accuracy in specific settings compared to other unpaired methods.
We focus on unpaired algorithms, particularly for deblurring and super-resolution tasks, where $\cA$ typically corresponds to a blur kernel. 
%Solving inverse problems without knowledge of the forward operator is a complex task which can be approached from different points of view. In \cref{tbl:setting} we provide a guide to disentangle the different settings: unlike many methods we will describe below, our algorithm only allows for a fixed degradation $\cA$ (cannot depend on the measurement $y$). On the other hand, by focusing on a specific degradation, it outperforms other unpaired methods in terms of accuracy. We focus here on unpaired algorithms, and mostly look at deblurring and super-resolution tasks (i.e.~$\cA$ is a blur kernel).
Data augmentation-based approaches~\cite{wang21realesrgan,zhang21bsrgan} construct synthetic supervised datasets using heuristically designed degradation pipelines which mimic diverse real-world corruptions. To improve adaptability to specific degradations,~\citet{zhang23sr} propose learning some of the pipeline parameters from a small dataset of noisy reference images.
%Data augmentation based approaches~\cite{wang21realesrgan,zhang21bsrgan} heuristically design a data-degradation pipeline which captures as many real-world cases as possible and use it to create a supervised dataset. 
%To improve the adaptivity to specific degradations~\citet{zhang23sr} learns some of the pipeline's parameters using a small reference noisy dataset. 
For better alignment with the degradation of each individual image, a line of work tackles the blind MAP inference problem using deep priors over both clean images and degradation operators. These methods are typically coupled with test-time optimization algorithms such as alternating minimization~\cite{ren20neural}, expectation-maximization (EM)~\cite{laroche2024fast,gao21deepgem}, or proximal gradient methods~\cite{vidal20sapg}.
%To conform even better to the degradation on each single image,  a class of algorithms consider the blind MAP problem using deep priors on both clean images and degradation operators, coupled with a test-time optimization algorithm such as alternating optimization~\cite{ren20neural}, EM~\cite{laroche2024fast,gao21deepgem}, or proximal gradient~\cite{vidal20sapg}. 
Recent approaches~\cite{chung2023parallel,laroche2024fast,yash24kerneldiff} incorporate diffusion models as priors for the blur kernel, jointly estimating $\cA$ and $x$ via plug-and-play algorithms~\cite{pnp}. Closely related methods include FKP~\cite{liang2021flow}, which uses a pretrained normalizing flow as a kernel prior, and DKP~\cite{yang2024dynamic}, which employs MCMC to iteratively refine the blur estimate. GibbsDDRM~\cite{murata2023gibbsddrm}, by contrast, uses a diffusion prior on the clean image and a simpler total variation (TV) prior on the blur kernel.
%A few recent methods~\cite{chung2023parallel,laroche2024fast,yash24kerneldiff} use a diffusion model as prior for the blur kernel and estimate $\cA, x$ with a plug and play algorithm~\cite{pnp}. 
%Closely related are FKP~\cite{liang2021flow} which also uses a pretrained kernel prior based on normalizing flows and DKP~\cite{yang2024dynamic} which uses MCMC to iteratively refine the blur estimate. GibbsDDRM~\cite{murata2023gibbsddrm} also uses a diffusion prior on clean images but a simpler TV prior on the blur kernel.

%In contrast to this class of methods which only need a single degraded image as input, what we propose does not require pretrained priors either on clean images or on the degradation operator, and can adapt precisely any noisy dataset on which it is trained. While it requires a small dataset from the same corruption process for training, its adaptation properties lead to better reconstruction accuracy.
In contrast to this class of methods, which require only a single degraded image at test time, our approach does not rely on any pretrained priors---neither on clean images nor on the degradation operator. Instead, it learns to adapt directly to the specific corruption process represented in the training data. While it does require a small dataset of degraded images from the same corruption distribution, this targeted adaptation enables significantly improved reconstruction accuracy.

Domain transfer approaches based on variations of the cycle-consistency loss~\cite{CycleGAN2017} are conceptually closer to our method. Given a noisy image $y$ and an unpaired clean image $x$, a clean-image generator $\cG$ and a noisy-image generator $\cF$ are jointly trained under the constraint that $\cG(\cF(x)) \approx x$ and $\cF(\cG(y)) \approx y$. For instance, CinCGAN~\cite{yuan18cincgan,maeda20unpaired} translates images downsampled with an unknown kernel into bicubically downsampled images, which can then be more effectively upscaled using standard super-resolution models.
Several related methods~\cite{bulat18imgsr,wolf21deflow,lugmayr19unsup,luo22pdmsr,sim20otcyclegan,mukherjee21UAR} employ generative models to synthesize corrupted images from clean ones, thereby enabling the construction of supervised training datasets. This synthesis can be done in a two-stage pipeline or directly in an end-to-end manner~\cite{romero22pcresrgan,chen20unsup}. Notably,~\citet{sim20otcyclegan} address a setting similar to ours, where a model is trained to deblur microscopy images in an unpaired setup.
Compared to these approaches, our method leverages a novel loss function derived from diffusion models—a technique not previously applied in this context. Unlike adversarial losses, our formulation is more stable and easier to train. Moreover, by learning an explicit representation of the degradation kernel rather than an implicit one, our method offers significantly improved interpretability. Furthermore, under certain assumptions, we are able to prove the identifiability of the true operator (see \cref{sec:method}).

%Domain transfer approaches based on variations of the cycle-consistency loss~\cite{CycleGAN2017} are more similar in spirit to our approach: given a noisy sample $y$ and an unpaired clean sample $x$, clean-image generator $\cG$ and noisy image generator $\cF$ are trained to obey the constraint that $\cG(\cF(x)) \approx x$ and $\cF(\cG(y)) \approx y$. CinCGAN~\cite{yuan18cincgan,maeda20unpaired} translate images downscaled with an unknown kernel into images downscaled with a bicubic kernel -- which can then be upscaled more easily; There exist many approaches~\cite{bulat18imgsr, wolf21deflow, lugmayr19unsup, luo22pdmsr, sim20otcyclegan, mukherjee21UAR} where generative models are used to synthesize corrupted images given clean ones such that a supervised training set can be easily created. This can be done in two stages or even in a single end-to-end step~\cite{romero22pcresrgan,chen20unsup}. In particular, \citet{sim20otcyclegan} includes a setting which is close to ours where it learns to deblur microscopy images.
%Compared to these methods we used a different loss based on diffusion models which has not yet been applied to this domain, and is not adversarial thus easier to train. By learning an explicit representation of the kernel instead of an implicit one we also provide much better interpretability. Furthermore, under certain assumptions, we are able to prove the identifiability of the true operator (see \cref{sec:method}).

We also briefly note that many methods aim to learn the degradation operator in a single-image fashion, but still require paired training data—typically synthetic—for supervision. For instance, IKC~\cite{gu2019blind} iteratively refines kernel estimates in alternation with clean-image predictions. DAN~\cite{luo23dan} extends this idea by unrolling the refinement process into an end-to-end trainable network. In contrast, KernelGAN~\cite{kligler19kernelgan} learns the degradation operator directly from a single input image, followed by a separate non-blind model to reconstruct the clean image $x$.

\subsection{Distribution matching}

Our method minimizes the distance between two probability distributions over images: the distribution of the observed noisy data $p(y)$ and that of clean data corrupted by a learned degradation operator, i.e., $p(\cA(x))$. A common strategy for such distribution matching is to use generative adversarial networks (GANs), which optimize an adversarial loss. GANs have been extensively applied to image restoration tasks, as discussed in the previous section~\cite{sim20otcyclegan,luo22pdmsr}, and can simultaneously learn both the degradation operator and its inverse within a unified training framework.
%Our method minimizes the distance between two probability distributions on images: the one of the true noisy data and the one of the clean data corrupted by the learned degradation operator (i.e. $p(y)$ and $p(\cA(x))$). Distribution matching can be done using GANs which minimize an adversarial loss. 
%GANs have been widely used in image restoration as detailed in the previous section~\cite{sim20otcyclegan,luo22pdmsr} and can learn not just the degradation operator but also its inverse in a unified training procedure. 
However, training GANs is notoriously challenging due to instability and sensitivity to hyperparameters~\cite{brock2018large}. Alternative approaches include normalizing flows, which provide exact likelihoods and invertible mappings. These have been used by DeFlow~\cite{wolf21deflow} for matching clean and noisy distributions, and by FKP~\cite{liang2021flow} to generate plausible blur kernels from single images.
%However, training such models is complex due to their adversarial nature, and there are many pitfalls to be aware of~\cite{brock2018large}. Alternatives include normalizing flows, used by DeFlow~\cite{wolf21deflow} to match clean and noisy distributions and by FKP~\cite{liang2021flow} to generate plausible blur kernels from single images.

More recently, diffusion models -- like GANs and normalizing flows -- have emerged as powerful tools for modeling empirical distributions and have been used in a range of distribution-matching tasks. For example, DreamFusion~\cite{poole22dreamfusion} learns 3D representations whose 2D projections are consistent with a pretrained diffusion model, while DiffInstruct~\cite{luo23diffinstruct} trains a single-step generator to match the distribution of a multi-step diffusion model. Both approaches rely on a loss function that approximates the KL divergence integrated over all diffusion time steps.

In our work, we use conditional flow matching (CFM) models~\cite{liu2022rectified,lipman23flowmatch} as a more conceptually straightforward alternative to standard diffusion models. We adapt the integrated KL divergence loss to the CFM framework for learning the degradation operator.
%More recently diffusion models -- which similarly to GANs and normalizing flows provide model-based representations of empirical distributions -- have been used for different distribution-matching tasks: in DreamFusion~\cite{poole22dreamfusion} a 3D representation was learned to such that its 2D projections could be consistent with a base diffusion model and in DiffInstruct~\cite{luo23diffinstruct} the goal was to learn a single-step generator with the same distribution as the multi-step base diffusion model. Both methods used the same loss, which can be seen as a KL-divergence integrated over all diffusion time-steps.
%We will be using conditional flow matching (CFM) models~\cite{liu2022rectified,lipman23flowmatch} instead of standard diffusion since they are conceptually cleaner, and will adapt the integrated KL divergence loss to such models.% We will occasionally use the terms diffusion and CFM interchangeably since at a high level they do not differ significantly.
\subsection{Camera lens calibration}
After preliminary experiments on synthetic deblurring, we focus on non-uniform deblurring in the context of a camera calibration pipeline. Camera calibration is a multifaceted task typically broken down into subtasks such as distortion, chromatic aberration, and vignetting corrections among others. In this work, we concentrate solely on compensating for blur induced by lens imperfections, with the goal of obtaining maximally sharp images.
%After preliminary synthetic deblurring experiments, we will focus on non-uniform deblurring as part of a camera calibration pipeline.
%Camera calibration is a complex task which is usually decomposed into many smaller problems: denoising, demosaicking, distortions, color aberrations, lens aberrations all need to be accounted for in order to clean up raw images. Here we are only focusing on correcting the blur introduced by lens defects in order to have maximally sharp results. 
These lens aberrations can be characterized by the point spread function (PSF) of the lens-camera system: the blur kernel that transforms an ideal point light source into a spread of colored spots in the image. PSFs are often non-uniform across the image plane and differ across RGB channels, resulting in chromatic aberrations. Correcting such distortions computationally is particularly appealing, as it enables the use of lower-cost lenses without sacrificing image quality.
%Such lens aberrations can be characterized by the point spread function (PSF) of the lens-camera system: the blur kernel which transforms an ideal point source of light into colored spots in an image.
%Typical PSFs are non-uniform over the image plane and vary across the RGB channels producing chromatic aberrations.
%Correcting for such defects computationally is attractive because: it allows cheaper lenses take better photographs; however, non-blind methods rely on knowing the PSF which can only be estimated through laborious procedures with complex setups~\cite{bauer2018automatic, kee2011modeling} involving specially printed patterns or screens. 

Traditional non-blind correction methods require precise knowledge of the PSF, which can only be obtained through laborious procedures involving printed calibration patterns or specialized screens~\cite{bauer2018automatic, kee2011modeling}. In contrast, blind lens aberration correction remains an understudied problem, although it shares many similarities with non-uniform deblurring. It is more commonly approached under the umbrella of \emph{defocus deblurring}~\cite{abuolaim2022improving, zamir21restormer, Lee2021IFAN, quan2023single}, where blur magnitude varies with scene depth. In aberration correction, however, the blur is dependent on spatial location in the image plane rather than depth.

%Blind lens aberration correction is certainly an understudied task, however it has many similarities with non-uniform deblurring. 
%More commonly such task is framed as \emph{defocus deblurring}~\cite{abuolaim2022improving, zamir21restormer, Lee2021IFAN,quan2023single} where the amount of blur depends on the depth of the imaged objects, contrary to aberration correction where the amount of blur depends on the image plane location.
Nonetheless, defocus deblurring methods may still be partially effective for lens aberration correction, especially when the induced blur is close to isotropic. As a starting point, we use the PSF dataset from~\citet{bauer2018automatic}, which provides ground-truth PSFs, and subsequently transition to a realistic calibration scenario using a Panasonic Micro 4/3 camera. In this setting, we acquire a small set of images and aim to learn the lens PSFs without access to ground-truth measurements.
%In any case defocus deblurring methods should be able to at least somewhat adapt to lens aberration when the true blur is approximately isotropic.
%We will initially use the PSF dataset collected in \citet{bauer2018automatic} to have access to a ground-truth, and then switch to a realistic calibration scenario where we take a few images using a Panasonic DSLR camera and learn its PSFs without access to the ground-truth.
%The main ad-hoc blind method we will compare with~\cite{eboli2022fast} uses a two-stage approach with deblurring followed by color-fringe removal~\cite{chang13tca,Eboli2023FastCA}.
Our main comparison will be with the blind method proposed by~\citet{eboli2022fast}, which performs deblurring followed by color-fringe removal in a two-stage pipeline~\cite{chang13tca,Eboli2023FastCA}.

    \section{Method}\label{sec:method}

Given an inverse problem of the form $y = \fwopt(x) + \epsilon$, with known noise level $\sigma$ and an unknown forward operator $\fwopt$ parameterized by a vector $\wopt$, we propose an algorithm to estimate $\wopt$ using unpaired data. On the one hand, we assume access to a clean dataset $\cX = \{ x_i \}_{i=1}^n$ of images drawn from a distribution $\dX$. On the other hand, we have access to a corrupted dataset $\cY = \{ y_j \}_{j=1}^m$ of images, generated from unknown clean samples via the forward model~$\fwopt$.
We denote by $\dYopt$ the distribution of such corrupted images, and by $\dYw$ the distribution induced by applying an arbitrary forward operator $\fw$ (with additive noise) to samples from $\dX$.

%Given an inverse problem of the form $y = \fwopt(x) + \epsilon$, with known noise level $\sigma$ and unknown forward operator $\fwopt$ parameterized by a vector $\wopt$, we propose an algorithm to learn $\wopt$ using unpaired data. One the one hand, we assume to have access to a clean dataset $\cX = \{ x_i \}_{i=1}^n$ of images drawn from a distribution $\dX$.
%On the other hand, we also have access to a corrupted dataset $\cY = \{ y_j \}_{j=1}^m$ of images generated from some unknown clean images via the forward model~$\fwopt$. We denote by $\dYopt$ the distribution of such images, and by $\dYw$ the distribution of images corrupted by some other operator $\fw$ (and additive noise).

Noting that $\dYopt$ is empirically accessible through $\cY$ while $\dYw$ can be approximated via $\cX$ and a candidate forward operator $\fw$, we hypothesize that minimizing the distance between $\dYopt$ and $\dYw$ with respect to $\fw$ yields a good approximation of the true forward model parameters:
% We hypothesize that minimizing the distance between $\dYopt$ --- empirically accessible through~$\cY$ —and $\dYw$ --- which can be approximated via $\cX$ and a candidate forward operator $\fw$—yields a good approximation of the true forward model parameters:
\begin{equation}
	\what = \argmin_{\whp} \cD(\dYopt, \dYw) \implies \fwhat \approx \fwopt,
\end{equation}
where $\cD$ is a distance between distribution that will be specified in detail later. We prove this rigorously in a simplified setting in \Cref{r:equiv}, with the full proof provided in the supplementary material, and we assume that the result generalizes when the distributions are only approximately equal, as is typically the case in practice.

%We show this rigorously in a restricted case in \cref{r:equiv}, the proof of which is in the supplementary material, and hypothesize that it holds when the distributions are only approximately equal as will happen in practice.
\begin{tprop}{}{equiv}
	For any set of forward model parameters $\whp$, let $p_\whp(y) = \int p_\whp(y\mid x) p(x) dx$ where $p_\whp(y\mid x) = \cN(y\mid A_\whp x, \sigma^2 I)$. Let $\wopt$ be a specific set of parameters which we consider to be the optimal set.
	Then, assuming the data covariance $\Sigma = \bE_x[xx^\top]$ is invertible, there exists an orthogonal matrix $P$ such that
	\begin{equation}
		p_{\whp}(y) = p_{\wopt}(y) \implies \fw = \fwopt\Sigma^{1/2}P\Sigma^{-1/2}
	\end{equation}
	That is, if the probability distributions $p_\whp$ and $p_{\wopt}$ are equal, it is possible to identify $\fwopt$ up to rotations $P$.
%	For any forward model parameters $\whp$, let $p_{\dYw}(y) = \int p_{\dYw}(y\mid x) p_{\dX}(x) dx$ where $p_{\dYw}(y\mid x) = \cN(y\mid \fw x, \sigma^2 I)$. 
%	Let $\wopt$ identify the optimal forward model $\fwopt$.
%	Then for every function $f$ of the form $f(x) = x^\top C x$ for any full-rank matrix $C$, there exists an orthonormal matrix $P$ such that
%	\begin{equation}
%		\bE_{y\sim \dYw, y'\sim \dYopt} [ f(y) - f(y') ] = 0 \implies P\fw - \fwopt = 0.
%	\end{equation}
%	Thus if the probability distributions $\dYw$ and $\dYopt$ are equal, under assumptions on $f$, it is possible to identify $\wopt$ up to rotations $P$.
\end{tprop}
Since neither $\dYopt$ nor $\dYw$ are explicitly available we will use CFM models to act as tractable representations of probability distributions on images.

\subsection{Distribution matching with diffusion}

A CFM model $v_\theta(z^{(t)}, t)$ trained on data from a distribution $\dYw$ allows one to sample from $\dYw$ by solving an ODE between times 0 and 1, defined as 
\begin{equation*}\label{eq:cfm_ode}
	dz^{(t)} = v_\theta(z^{(t)}, t) dt, \quad z^{(0)} \sim \cN(0, I). %\text{ s.t.} z^{(t)} \sim (1 - t) \cN(0, I) + t
\end{equation*}
such that $z^{(t=1)} \sim \dYw$. 
This conditional flow matching~\cite{liu2022rectified, lipman23flowmatch} perspective is connected to standard Langevin diffusion since the velocity field $v_\theta(x^{(t)}, t)$ is related to  the \emph{score} of a diffusion model: $v_\theta(z^{(t)}, t) + z^{(t)}\propto\nabla_z\log p_{\dY[t]{\whp}}(z^{(t)}_{\whp})$~\cite{zhang2024flow}, assuming the neural network model of the velocity field to be exact. 
Each $z^{(t)}$ follows an intermediate distribution between Gaussian noise and the data $\dY[t]{\whp}$ for $t\in[0, 1]$.
Hence, instead of computing a distance between two distributions, we will compute it between two sequences of distributions $\dY[t]{\whp}$ and $\dY[t]{\wopt}$ which arise from two CFM models. In particular we will use the KL divergence integrated over time, and follow the derivation proposed by DiffInstruct (DI)~\cite{luo23diffinstruct} to compute its gradient with respect to $\whp$.
The integrated KL divergence is defined as
\begin{equation}\label{eq:ikl}
	\IKL{\dYopt}{\dYw} = \int_{t=0}^1 \kld{\dY[t]{\wopt}}{\dY[t]{\whp}} dt.
\end{equation}
Assuming that the score terms for both probability distributions (i.e. $s_{\theta,\wopt}(y,t) \approx \nabla_y \log p_{\dY[t]{\wopt}}(y)$ and $s_{\phi,\whp}(y,t) \approx \nabla_y \log p_{\dY[t]{\whp}}(y)$) can be computed, it is possible to efficiently differentiate the IKL with respect to $\whp$:
\begin{equation}\label{eq:di-grad}
\begin{aligned}
	\nabla_{\whp} \IKL{\dYopt}{\dYw} = 
	\int_{t=0}^1 \bE \big[ 
	s_{\theta, \wopt}(y^{(t)}_{\whp}, t) - s_{\phi, \whp}(y^{(t)}_{\whp}, t) \big]^\top \nabla_{\whp} y^{(t)}_{\whp} dt,
%	\nabla_y \log p_{\dY[t]{\wopt}}(y^{(t)}_{\whp}) - \nabla_y \log p_{\dY[t]{\whp}}(y^{(t)}_{\whp}) \big]^\top \nabla_{\whp} y^{(t)}_{\whp} dt
	% _{\substack{y_{\whp}^{(0)} \sim \cN(0, I), x \sim \cX \\ y_{\whp}^{(1)} = \fw(x)+\epsilon }}
\end{aligned}
\end{equation}
%\begin{equation}\label{eq:di-grad}
%	\nabla_{\whp} \IKL{\dYopt}{\dYw} = \int_{t=0}^1 \bE_{\substack{y_{\whp}^{(0)} \sim \cN(0, I), x \sim \cX \\ y_{\whp}^{(1)} = \fw(x)+\epsilon }} \big[ \nabla_y \log p_{\dY[t]{\wopt}}(y^{(t)}_{\whp}) - \nabla_y \log p_{\dY[t]{\whp}}(y^{(t)}_{\whp}) \big]^\top \nabla_{\whp} y^{(t)}_{\whp} dt
%\end{equation}
where the expectation is over $y_{\whp}^{(0)} \sim \cN(0, I), x \sim \cX$, $y_{\whp}^{(1)} = \fw(x)+\epsilon$ and  $y^{(t)}_{\whp} = (1 - t) y^{(0)}_{\whp} + t y^{(1)}_{\whp}$. Note that \cref{eq:di-grad} has in fact been originally introduced to optimize a 3D scene consistent with a diffusion model's outputs~\cite{poole22dreamfusion} and in \cite{luo23diffinstruct} was used to learn a distilled model. Here we are showing it can be used effectively on conditional flow matching models as well, and in a completely different setting.
The loss gradient shown in \cref{eq:di-grad} needs certain quantities to be computed: the score of $p_{\dYopt}$ can be obtained beforehand by training a flow-matching model $v_{\theta, \wopt}$ on the available noisy data $\cY$. The score of $p_{\dYw}$ however depends on a distribution which changes with every change in $\whp$. Therefore the strategy we use is to alternately optimize \begin{enumerate*}[label=\roman*)]\item an \emph{auxiliary flow-matching model} $v_{\psi,\whp}$ for a fixed $\whp$ with a standard diffusion loss and \item forward model parameters $\whp$ with the IKL loss keeping the auxiliary model fixed~\cite{luo23diffinstruct}\end{enumerate*}.
By changing the parametrization of $\fw$ we can use \cref{eq:di-grad} to learn the degradation of a variety of different inverse problems. We will now go into the details of the algorithm for the task of learning blur operators.

\subsection{Learning the forward operator}

In deblurring, $\fw$ corresponds to a convolution with kernel $\whp$. The algorithm we propose proceeds in three distinct steps: \begin{enumerate*}[label=\roman*)] 
	\item learn the corrupted data distribution, 
	\item approximate the forward operator $\fwopt$ and \item solve the non-blind inverse problem.
\end{enumerate*}

\begin{figure}
	\centering
	\begin{minipage}{0.57\linewidth}
		\includegraphics[width=0.95\linewidth]{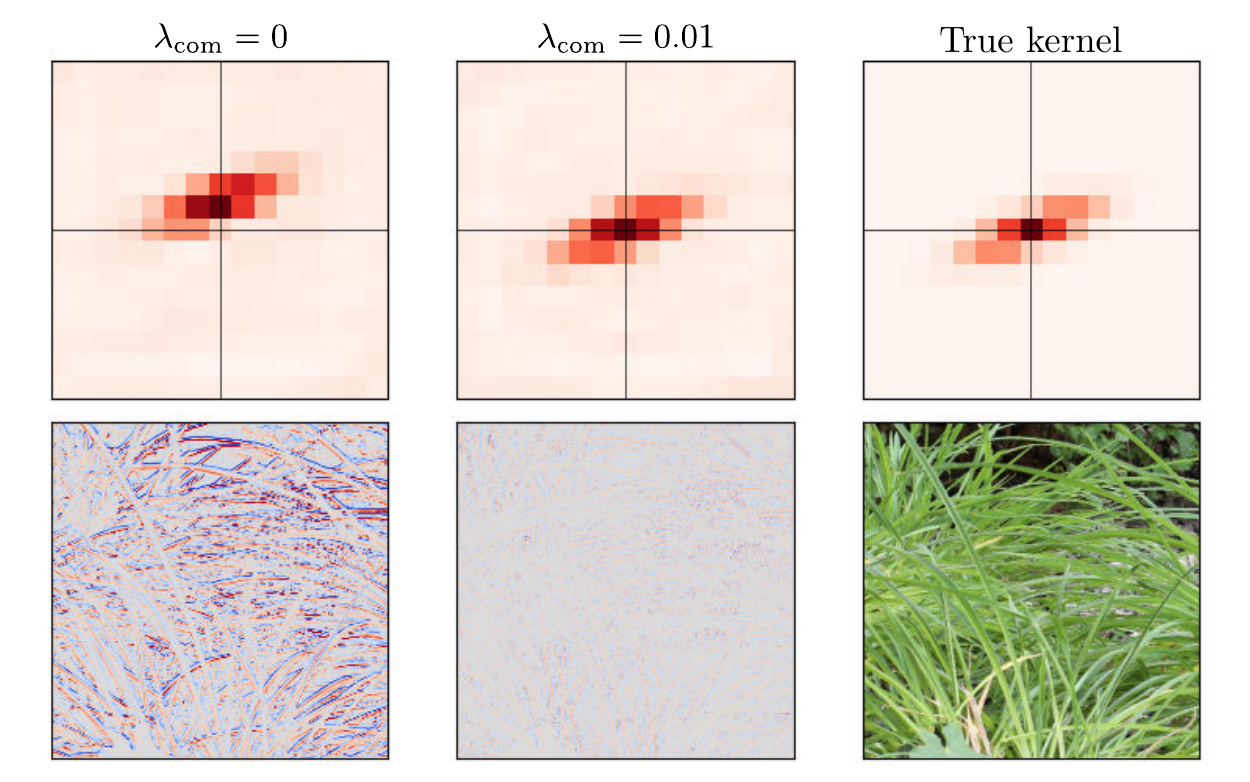}
		\caption{Effect of center regularization on reconstruction quality. Without regularization the learned blur filter may be shifted leading to larger reconstruction errors (second row of the plot). A moderate amount of regularization fixes this.}
		\label{fig:center-reg}
	\end{minipage}\hfill{}
	\begin{minipage}{0.35\linewidth}
		\centering
		\includegraphics[width=0.95\linewidth]{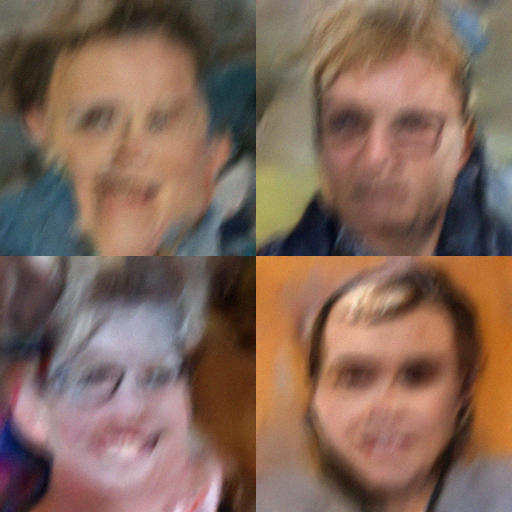}
		\caption{Samples generated by flow matching model with motion-blur kernel. While the model is average at generating faces, it correctly represents the degradation.}
		\label{fig:ffhq-deg}
%		\centering
%		\includegraphics[width=0.95\linewidth]{img/imggen_cond.png}
%		\caption{Images generated by a location-conditional model with the same initial noise and varying conditioning. The model disentangles content and degradation.}
%		\label{fig:noisy_patch}
	\end{minipage}
\end{figure}

\paragraph{Step 1: Learning $\dYopt$}
In more detail, the first step uses the noisy dataset $\{y_j\}_{j=1}^m$ to train a conditional flow matching (CFM) model $v_{\theta, \wopt}$ that transports Gaussian noise samples at time $t=0$, $z^{(0)} \sim \mathcal{N}(0, I)$, to samples from the corrupted data distribution $\dYopt$ at time $t=1$, i.e., $z^{(1)} \sim \dYopt$. The training objective is the standard conditional flow matching loss~\cite{liu2022rectified,lipman23flowmatch}:
%In more detail, the first step consists of using noisy data $\{y_j\}_{j=1}^m$ to train a CFM model $v_{\theta, \wopt}$ to transport random Gaussian noise samples at time 0, $z^{(0)}\sim\cN(0, I)$ into samples from the distribution $\dYopt$ at time 1, $z^{(1)} \sim \dYopt$.
%The training loss for this model is the standard one for conditional flow matching~\cite{liu2022rectified,lipman23flowmatch}
\begin{equation*}
	\cL_{\mathrm{CFM}} = \mathbb{E}_{t, z^{(0)}, z^{(1)}} \big[ \norm{v_{\theta,\wopt}(z^{(t)}, t) - (z^{(1)} - z^{(0)}) } \big].
\end{equation*}
Importantly, for the overall success of our method, it is not necessary for $v_{\theta,\wopt}$ to achieve state-of-the-art generative quality; it only needs to effectively capture the degradation process, which we found to be significantly easier than precisely modeling image content. See \cref{fig:ffhq-deg} for sample outputs produced by a model at this stage.
%Note that for the success of the whole algorithm it is not necessary for $v_{\theta,\wopt}$ to be a state of the art generative model: it only needs to capture the degradation, which we found to be much easier than to precisely capture content (see \cref{fig:ffhq-deg} for samples generated by a model at this stage).
%\begin{figure}
%    \centering
%    \includegraphics[width=0.9\linewidth]{img/ffhq_mb_generated_crop.png}
%    \caption{Samples generated by flow matching model with motion-blur kernel. 32 steps were used for generation.}
%    \label{fig:ffhq-deg}
%\end{figure}
%Note that even though we learn the ODE's velocity $v_\theta$, in case the starting point of the ODE is a Gaussian distribution, $v_\theta$ is linearly\todo{not really but close, need to check the prop} related to the \emph{score function} $\nabla_{y_t}...$ \cite[Proposition~1]{zhang2024flow} and hence we have the information needed to implement~\cref{eq:di-grad}.

\paragraph{Step 2: distribution matching}
The second step uses the clean dataset $\cX$ to learn the corruption operator $\hatfw \approx \fwopt$ by minimizing the IKL loss~\eqref{eq:ikl}. This involves two alternating optimization steps: first, training an auxiliary diffusion model $v_{\phi,\whp}(z^{(t)}, t)$ using the standard flow-matching loss, where $z^{(t)} = (1-t)z^{(0)} + t \fw(x)$; second, updating $\hatfw$ by following the gradient in \cref{eq:di-grad}. 

To encourage fast convergence, the auxiliary model $v_{\phi,\whp}$ is initialized with the pretrained weights from $v_{\theta,\wopt}(z^{(t)}, t)$. The parameterization of $\hatfw$ is flexible: it may depend on the clean image $x$—which limits certain options for the final step—or be independent of $x$. For instance, we consider modeling non-uniform blur with a an operator $\hatfw$ that varies with pixel location. The framework is general, but $\hatfw$ should not depend on corrupted images, as these are unavailable in the unpaired training setting.

%The second step uses the clean dataset $\cX$, to learn the corruption operator $\hatfw\approx\fwopt$ by minimizing the IKL~\eqref{eq:ikl}. 
%The algorithm requires two alternating optimization steps: the first follows the standard flow-matching loss to train auxiliary diffusion model $v_{\phi,\whp}(z^{(t)}, t)$ with $z^{(t)} = (1-t)z^{(0)} + t \fw(x)$, the second follows the gradient in \cref{eq:di-grad} to learn $\hatfw$. 
%To encourage fast convergence, the auxiliary model should be initialized to the same weights as the pretrained model $v_{\theta,\wopt}(z^{(t)}, t)$ but $\hatfw$ can be parameterized in various ways: it can depend on the clean image $x$, in which case the options for the last step will be limited, or it can be independent. 
%One option we will consider is to model non-uniform blur with $\hatfw$ which depends on the pixel-location. The general framework is flexible but $\hatfw$ should not depend on the corrupted images, since these are not available in the unpaired training framework. 

The distribution matching step can also be regularized in various ways to introduce prior knowledge about the forward model and improve the quality of results. For example, when $\cY$ and $\cX$ consist of patch data, their distribution will be invariant to translation: an image patch translated by some amount in any direction will still follow the same distribution. This invariant can easily lead to learning blur filters which are off-center as shown in \cref{fig:center-reg}. To counter this we add a regularizer which constrains the center of mass of the learned kernels to be in the middle of the filter itself.
%\begin{equation}\label{eq:center-reg}
%	\Big(\frac{k_x}{2} - \mathrm{com}_x(\hatfw)\Big)^2 + \Big(\frac{k_y}{2} - \mathrm{com}_y(\hatfw)\Big)^2.
%\end{equation}

\paragraph{Step 3: solve the inverse problem}
The third and final step is to use the learned forward model $\hatfw$ to solve the inverse problem. At this point the non-blind setting applies, hence multiple strategies can be chosen depending on the specific problem and on the type of data which is available. 
When a larger clean-image dataset is available, by leveraging $\hatfw$ it is possible to generate a paired noisy-clean dataset on which to train a supervised image restoration model~\cite{wang18esrgan,zamir21restormer}. 
The second option is to use a plug and play algorithm~\cite{pnp} which leverages a pretrained prior on clean images (classically this could have been a prior such as total-variation instead) to iteratively convert a noisy image into a clean one. In \cref{sec:exp} we will use ESRGAN~\cite{wang18esrgan} from the first option and DPIR~\cite{zhang17dpir}, DiffPIR~\cite{zhu23diffpir} from the second one. Of course classical alternatives such as Wiener deconvolution can be used depending on the specific inverse problem.

	\section{Experiments}\label{sec:exp}
\subsection{Deblurring}
We used subsets of the FFHQ dataset~\cite{ffhq} to compare with blind deblurring methods. In particular, for each of two degradation operators we train a small CFM model on 1000 images from FFHQ corrupted by the blurring operator $\fwopt$ and subsequently utilize 100 different clean images to learn $\hatfw$ with distribution matching. We use an isotropic Gaussian blur with standard deviation \num{1} and a motion blur kernel generated following \citet{motionblur}. In both cases, Gaussian noise is added with standard deviation of \num{0.02}. 
We finally use plug and play algorithm DiffPIR~\cite{zhu23diffpir} to solve the inverse problem $\hatfw$. 
%Note that the diffusion model doesn't have to be particularly large or well trained, as it will learn the corruption distribution much earlier than the precise content distribution. Sample images from our FFHQ-trained diffusion model are shown in the supplementary material.
The natural upper baseline for this experiment is to run the same PnP algorithm with the ground-truth kernel. 
In addition to this, we compare with some lower baselines which learn the correct kernel from a \emph{single image}, and simultaneously perform deblurring: diffusion based methods BlindDPS~\cite{chung2023parallel}, FastDiffusionEM~\cite{laroche2024fast} and KernelDiff~\cite{yash24kerneldiff}. These latter blind algorithms require two priors (in the form of pretrained diffusion models): one on the image dataset and one on the kernels. BlindDPS and FastDiffusionEM use FFHQ as image prior while KernelDiff uses several natural image datasets~\cite{dong2020deep}. All three methods assume the blur kernels to be motion-blur, generated using the same stochastic procedure~\cite{motionblur} as the true one. BlindDPS additionally includes isotropic Gaussian kernels in its kernel prior, which is why in \cref{tbl:ffhq} we also test it on a Gaussian kernel. 
It is important to stress that our algorithm works in a different setting from the single-image methods: we require a dataset of clean and noisy images for each degradation, while the single-image algorithms only require a single noisy image. However, note that all single-image algorithms rely strongly on the image and kernel priors on which they were trained: for example we could not successfully run the method from~\citet{laroche2024fast} on a simple Gaussian kernel without first retraining the kernel prior, and KernelDiff which uses a different image prior than FFHQ performs poorly in this setting.
The results in \cref{tbl:ffhq} demonstrate two things: first, even when the problem is purely \emph{in-distribution}, the gap between blind and non-blind algorithms is large. In second instance, the algorithm we propose significantly reduces this gap by using more data from the same distribution to learn the necessary information about the degradation operator.
To increase the robustness of our experiments we compute the standard deviation over 5 different random seeds for the 2nd step of our pipeline (thus keeping the 1st step fixed). For the motion blur kernel, we find a standard deviation of 0.02 for PSNR and 0.0005 for LPIPS which are both negligible.

\begin{table}
	\centering
	\caption{Reconstruction error on FFHQ. KernelDiff was run with no noise. FastEM was run with 16 samples and $\Pi$GDM.}
	\label{tbl:ffhq}
	\begin{tabular}{lllll}
		\toprule
		& \multicolumn{2}{c}{Gaussian ($\sigma=1$)} & \multicolumn{2}{c}{Motion} \\
		& \multicolumn{2}{c}{\includegraphics[width=0.1\linewidth]{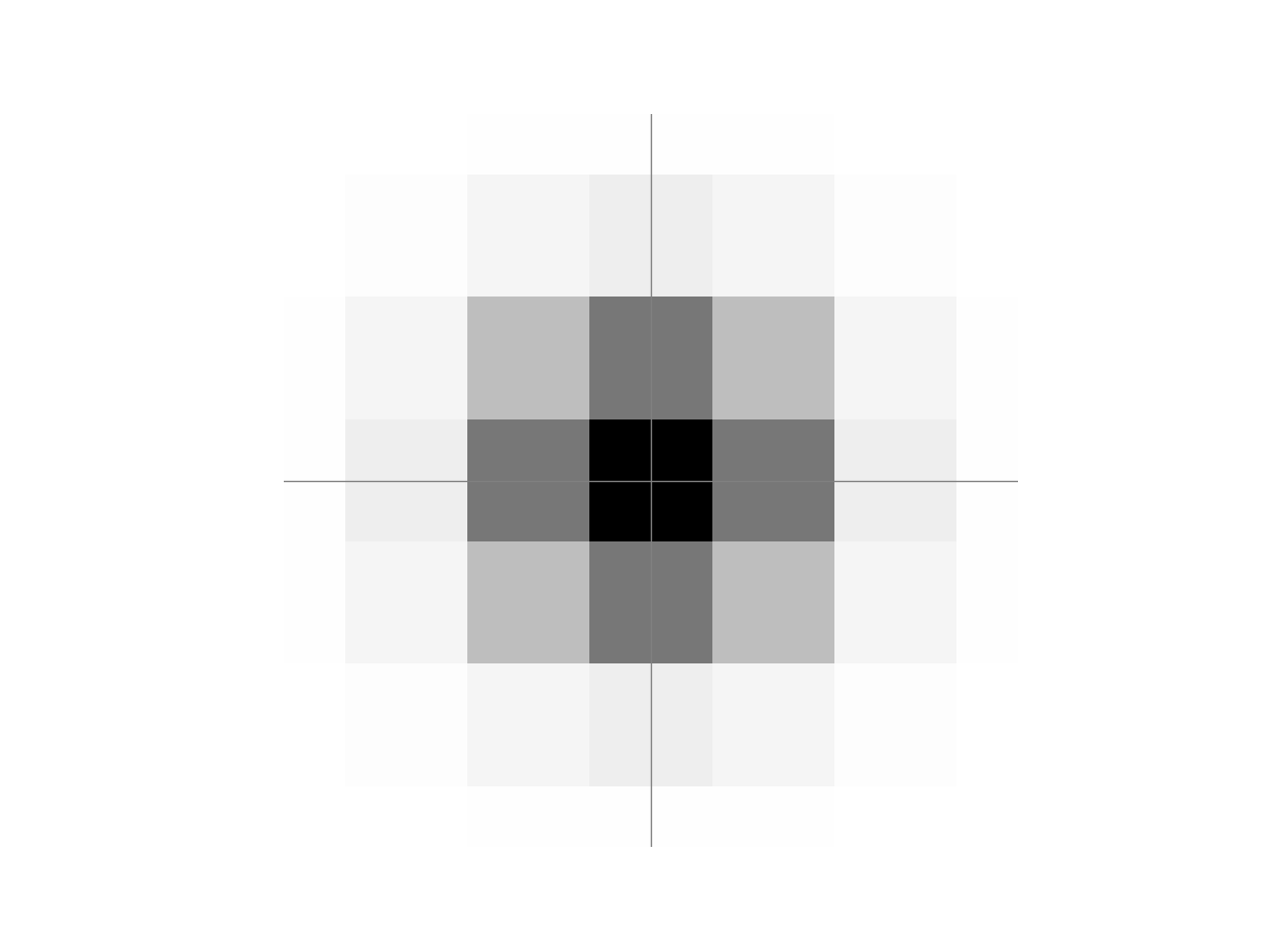}} & \multicolumn{2}{c}{\includegraphics[width=0.1\linewidth]{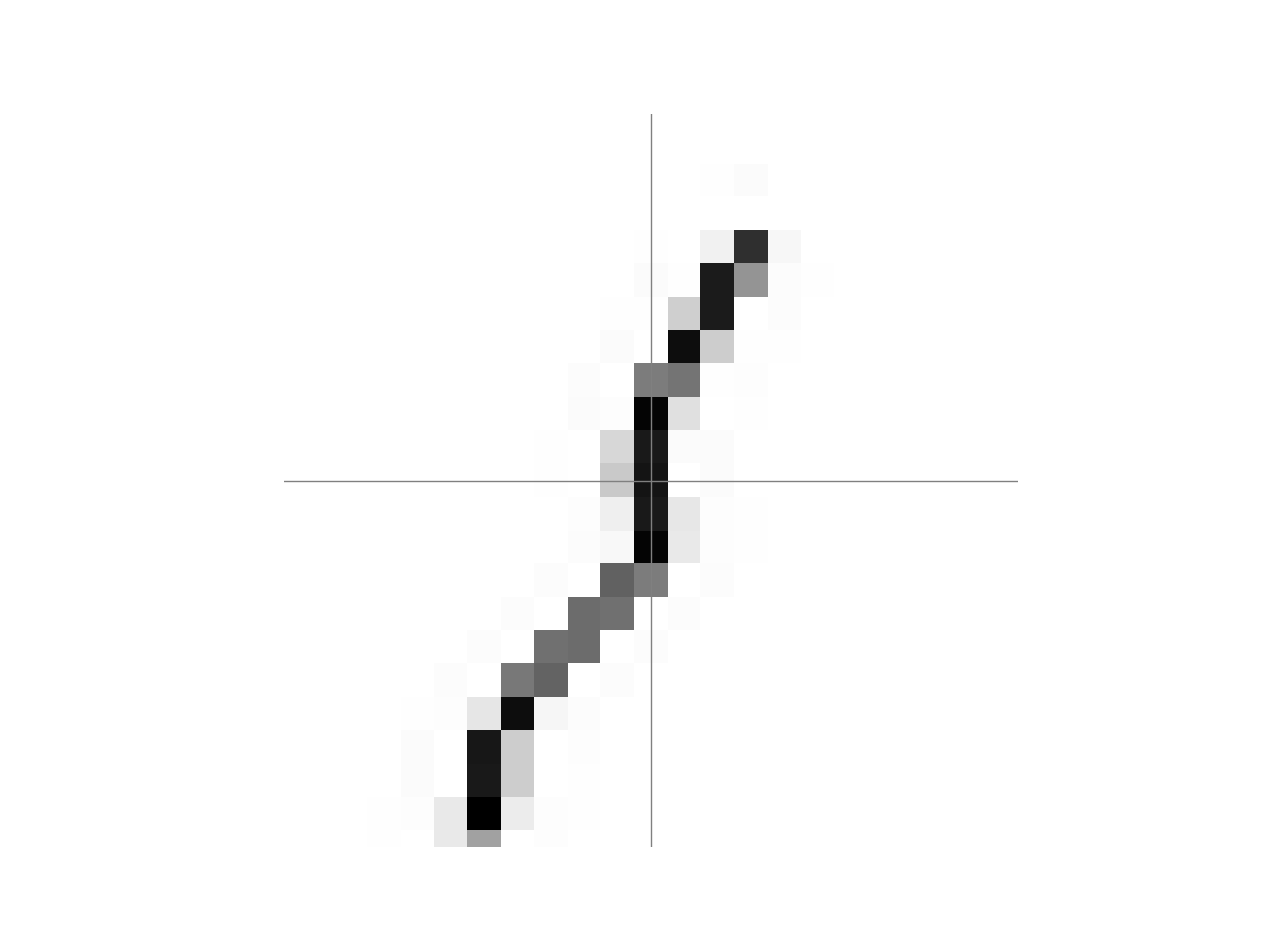}} \\
		& PSNR $\uparrow$ & LPIPS $\downarrow$ & PSNR $\uparrow$ & LPIPS $\downarrow$ \\
		\midrule 
		\multicolumn{5}{c}{\textit{Non-blind (true kernel available)}} \\
		DiffPIR                                      & 32.7 & 0.031 & 29.3 & 0.068 \\
		\midrule
		\multicolumn{5}{c}{\textit{Blind, multi-image}} \\
		Ours + DiffPIR                                  & 32.7 & 0.031 & 28.8 & 0.069 \\
		\midrule
		\multicolumn{5}{c}{\textit{Blind, single-image}} \\
		Blind-DPS~\cite{chung2023parallel}            & 29.8 & 0.077 & 25.5 & 0.122 \\
		FastEM (n=16)~\cite{laroche2024fast} & --   & --    & 24.9 & 0.130 \\
		KernelDiff~\cite{yash24kerneldiff}            & --   & --    & 21.7 & 0.194 \\
		% diy_tr1k_ffhq256-5000-6000_ksmb_ns0.02_lr1e-5x4_1is_sreg1_direct_v3_0.25Mdpir 
		%			Ours (1k-nobord)+DPIR                         & 31.4 & 0.122 & 29.5 & 0.231 \\
		%  diy_tr1k_ffhq256-5000-6000_ksmb_ns0.02_lr1e-5x4_1is_sreg1_direct_v3_0.25Mdiffpir 
%		\midrule
		%			DPIR*			 &      &       & 30.3 & 0.225\\
		\bottomrule
	\end{tabular}
\end{table}

\subsection{Space-varying blur}
In practice image blur may come from a variety of sources such as camera shake/motion, out-of-focus objects or lens distortions. 
We focus on the latter which better fits the our framework: it is easy to collect a set of noisy images with equal distribution, but it is not really possible to collect paired datasets (note that  paired datasets can be generated synthetically through software camera models~\cite{li2021universal}). 
Every imaging system is imperfect, which can be characterized by a spatially varying point spread function (PSF) at every location in the image plane. It can thus be modeled as a per-pixel blur where the blur kernel depends on the image plane location.
Since the PSFs are intrinsically connected to the imaging system, multiple pictures taken with the same camera and camera settings will have the same degradation.

\begin{figure}
\centering
%\subcaptionbox{True kernels} [0.32\linewidth]{\includegraphics[width=0.95\linewidth]{img/ddpd_kernel_err_1.png}}
%\subcaptionbox{Kernel error} [0.32\linewidth85{\includegraphics[width=0.95\linewidth]{img/ddpd_kernel_err_2.png}}
%\subcaptionbox{Reconstruction error} [0.32\linewidth]{\includegraphics[width=0.95\linewidth]{img/ddpd_kernel_err_3.png}}
\includegraphics[width=0.9\linewidth]{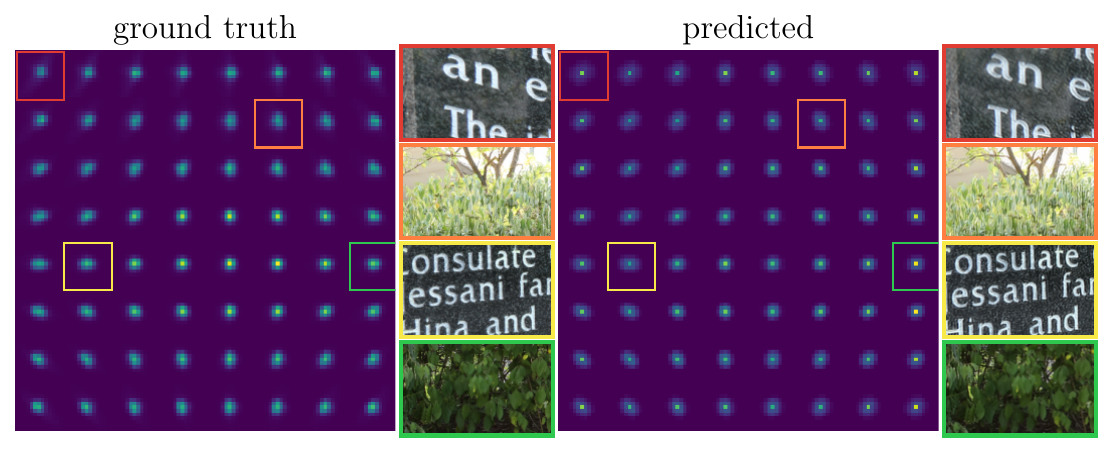}
\caption{Comparing ground-truth (left) and predicted blur kernels (right), as well as the respective reconstructed images.}
\label{fig:ddpd_kernel_err}
\end{figure}

For the first experiment we use real PSFs, taken from a subset of those identified in a real camera system~\cite{bauer2018automatic} but synthetically applied to a clean dataset. In particular we use the green channel PSFs from the Canon EF 24mm f/1.4L II USM lens at 1.4 aperture, and subsample them to a 8x8 grid, shown in \cref{fig:ddpd_kernel_err}~(top-left). This grid is then mapped to images from DIV2K~\cite{agustsson17div2k} and DPDD~\cite{abuolaim2020dpdd}, and applied as a per-pixel blur by linearly interpolating the kernels to each image location.
We use the degraded DIV2K training-set (subdivided into patches 128 pixels wide) to train the diffusion model in the first stage of our algorithm, and the clean DDPD training-set for the second stage -- thus ensuring the data is strictly \emph{unpaired}.
For the third and final stage we experiment with two procedures: the plug and play algorithm DPIR~\cite{zhang17dpir} which uses a CNN as regularizing prior and can be applied directly to the learned kernels, and supervised method ESRGAN~\cite{wang18esrgan} for which we generate a paired clean-noisy dataset using the learned degradation.
In order to condition the kernels on image location, we add two positional encoding channels to our images such that the diffusion model learns different distributions based on the patch location. For the degradation prediction we directly learn the 64 kernels without any additional parametrization; both at train and at test time the correct kernels are picked by using the positional encoding and linearly interpolating between the kernels.
For this experiment we used the centering regularization and also introduced an isotropic Gaussian regularization which helped stabilize training. 

\begin{table}
	\centering
	\caption{Reconstruction error for non-uniform deblurring on DPDD. The degradation is a spatially varying blur with additive Gaussian noise ($\sigma=0.01$).}
	\label{tbl:dpdd}
	\begin{tabular}{llll}
		\toprule
		Method & PSNR $\uparrow$ & SSIM $\uparrow$ & LPIPS $\downarrow$ \\
		\midrule
		\multicolumn{4}{c}{\textit{Non-blind (true kernel available)}} \\
		ESRGAN~\cite{wang18esrgan}       & 32.53 & 0.906 & 0.058 \\
		DPIR~\cite{zhang17dpir}          & 34.89 & 0.933 & 0.112 \\
		\midrule
		\multicolumn{4}{c}{\textit{Blind, multi-image}} \\
		DeFlow+ESRGAN~\cite{wolf21deflow} & 27.92 & 0.818 & 0.206 \\
		Ours+ESRGAN                       & 32.38 & 0.911 & 0.061 \\
		Ours+DPIR                         & 33.84 & 0.933 & 0.099 \\
		\midrule
		\multicolumn{4}{c}{\textit{Blind, single-image}} \\
		INIKNet~\cite{quan2023single}     & 29.45 & 0.874 & 0.160 \\
		Restormer~\cite{zamir21restormer} & 28.59 & 0.859 & 0.203 \\
		\bottomrule
	\end{tabular}
\end{table}

\begin{figure}
	\centering
	\subcaptionbox*{\footnotesize Full ours }[0.22\linewidth]{\includegraphics[width=0.99\linewidth]{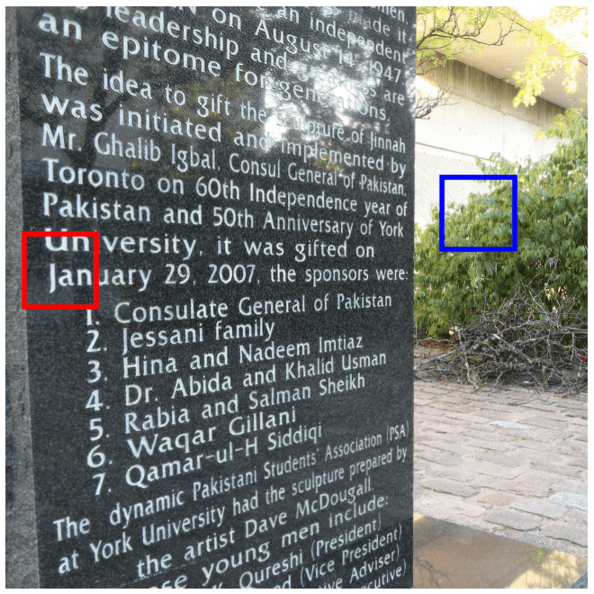}}
	\subcaptionbox*{\footnotesize Measurement }[0.123\linewidth]{\includegraphics[width=0.913\linewidth]{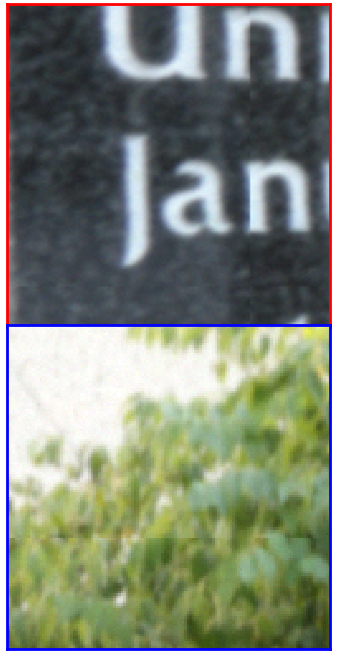}}
	\subcaptionbox*{\footnotesize INIKNet \cite{quan2023single} }[0.123\linewidth]{\includegraphics[width=0.913\linewidth]{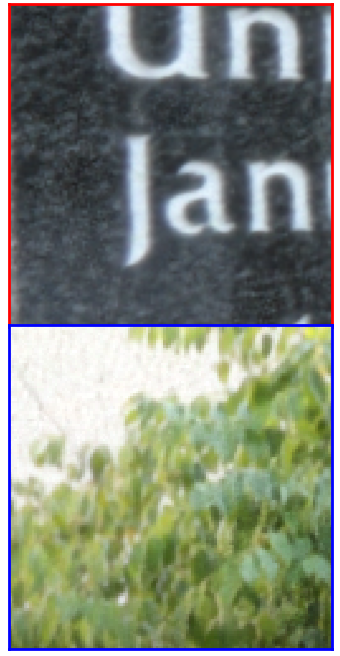}}
	\subcaptionbox*{{\footnotesize Restormer \cite{zamir21restormer}} }[0.123\linewidth]{\includegraphics[width=0.913\linewidth]{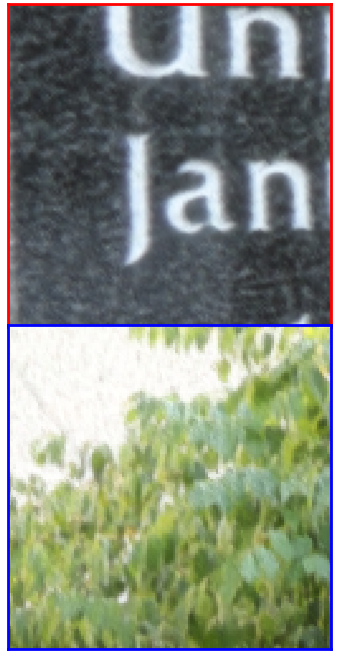}}
	\subcaptionbox*{\footnotesize DeFlow \cite{wolf21deflow} }[0.123\linewidth]{\includegraphics[width=0.913\linewidth]{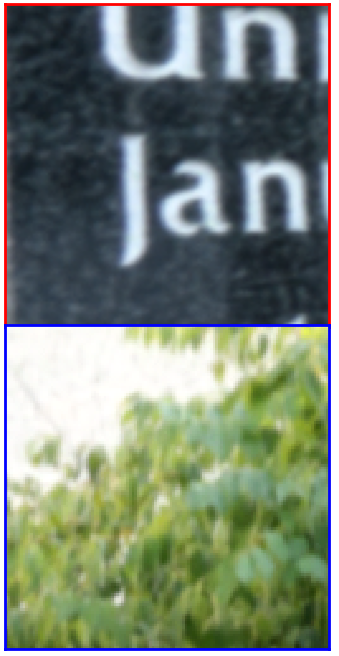}}
	%	\subcaptionbox{ESRGAN*}                [0.11875\linewidth]{\includegraphics[width=0.95\linewidth]{img/ddpd_comparison_esrgan.png}}
	\subcaptionbox*{\footnotesize Ours+DPIR }[0.123\linewidth]{\includegraphics[width=0.913\linewidth]{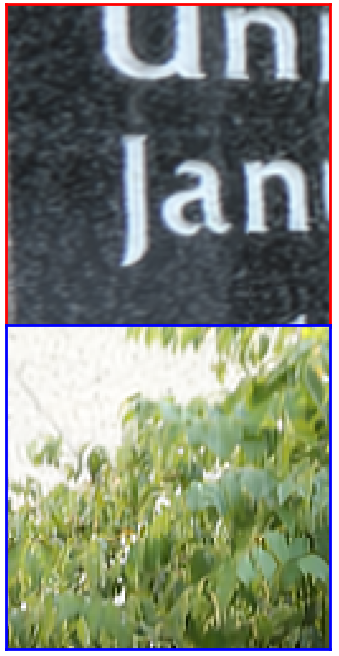}}
	\subcaptionbox*{\footnotesize DPIR*~\cite{zhang17dpir} }[0.123\linewidth]{\includegraphics[width=0.913\linewidth]{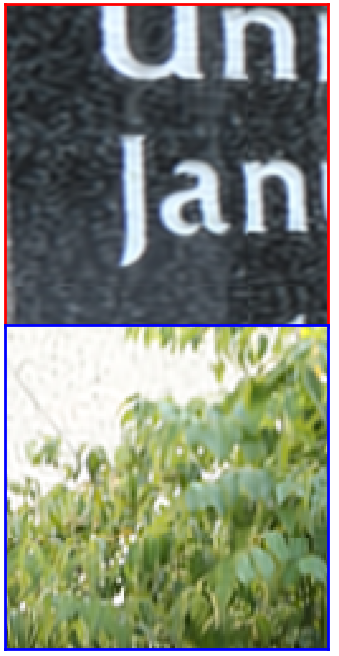}}
	\caption{Sample reconstructions on DDPD. INIKNet and Restormer are single image, DPIR is non-blind.}
	\label{fig:ddpd_comparison}
\end{figure}

We compare the results obtained on the ``target'' test-set of DPDD (corrupted with the camera PSFs and not with the original defocus degradations) against two pretrained methods tailored for real-world defocus deblurring: Restormer~\cite{zamir21restormer} and INIKNet~\cite{quan2023single}, against unpaired algorithm DeFlow~\cite{wolf21deflow} which we retrain for the current task and couple with ESRGAN (DeFlow does not provide explicit degradations hence cannot be coupled with DPIR) and against the upper baselines ESRGAN and DPIR trained with the true degradation.
While Restormer and INIKNet were trained on a different degradation domain, defocus blur is also spatially varying and should not be too far from the mostly isotropic kernels of the camera PSF. 
Nevertheless, their inferior performance compared to our method shows how even small changes to the degradation distribution can have a large impact on reconstruction performance.
\Cref{tbl:dpdd} shows the quantitative evaluation on DPDD while the qualitative results are shown in \cref{fig:ddpd_comparison}. DeFlow does not manage to learn the correct blur distribution, and simply introduces a small amount of noise in the generated \emph{degraded} images, thus garnering the worst results. 
Both INIKNet and Restormer perform similarly and succeed at removing some of the blur, but the results are not as sharp as the reconstructions obtained with our method. Both non-blind methods perform very well, with ESRGAN being better under perceptual metrics and DPIR under distortion metrics. Importantly, and like we showed for the first round of experiments on FFHQ, our method is very close to its upper limit given by the non-blind counterpart.
Note that, as shown in \cref{fig:ddpd_kernel_err}, the predicted kernels are not isotropic despite our regularizer and manage to capture the spread and directional variations of the true kernels. However, there remain a few kernels such as the one at the top-left whose long tails we cannot capture well. This leads to under-compensating the blur as can be seen in the left-most bottom panels of \cref{fig:ddpd_kernel_err}.
%To better understand the pixel-location conditioning, in \cref{fig:noisy_patch} we show some samples from the CFM model $s_{\theta, \wopt}$ (trained on noisy data) generated from the same initial noise but conditioned on different patch locations. Comparing them to the ground-truth kernels, the network is able to decouple the content of the generated patch which remains stable from the degradation which varies.

\subsection{Real-world camera lens calibration}
\begin{figure}
	\centering
	\includegraphics[width=0.75\linewidth]{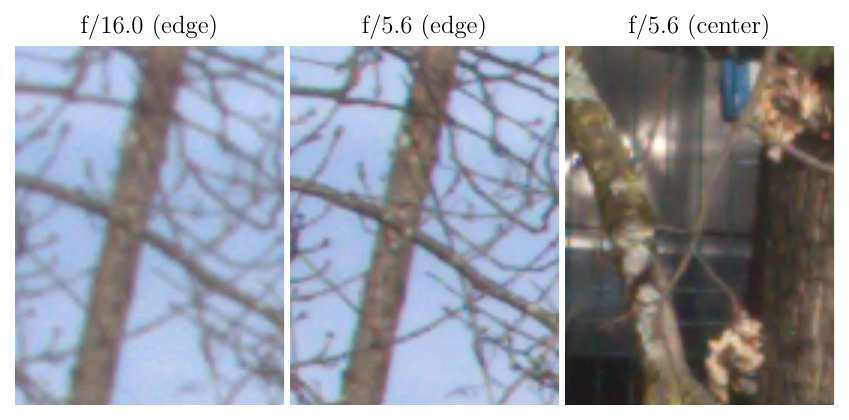}
	\caption{Different aberrations in parking lot data. While images taken at f/5.6 are sharp, chromatic aberrations are still present outside of the center portion as evidenced by the middle panel.}% middle shows zoom-ins of the same, lower panel shows the f/16 picture at the same location (noticeably more blurry).}
	\label{fig:p-lot-data}
\end{figure}
\begin{figure}
	\centering
	\includegraphics[width=0.97\linewidth]{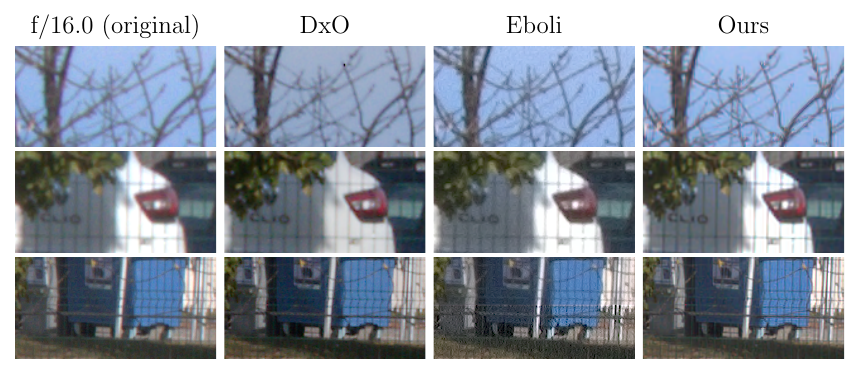}
	\caption{Sample reconstructions on the parking lot dataset. Color differences in DxO are likely due to a different white-balancing algorithm. \citet{Eboli2023FastCA} correctly removes chromatic aberrations but introduces some noise artifacts. Our method results visually pleasing and significantly sharper than the original image.}
	\label{fig:p-lot-recon}
\end{figure}
As a final experiment we attempt to tackle the same lens-aberrations of the previous task, but this time on real data. We used a Panasonic DC-GX9 camera in aperture priority mode with a Leica DG Summilux 25mm f/1.4 II lens to take 22 pictures in our parking lot at different apertures. Our goal was to learn the PSF of the lens at an extreme aperture (e.g.~f/16) in order to correct it using the clean distribution of images taken at a reasonable aperture (e.g.~f/5.6). 
To avoid confounding factors we tried to keep all the image-plane in focus and had ample light to obtain sharp images. 
Note that while the images at different f-stops were taken using a tripod from the same location no additional care was taken to align them, and they would not be suitable for a supervised learning algorithm.
Images were minimally postprocessed, by devignetting and converting to sRGB using lensfunpy~\cite{lensfunpy}, in order to preserve the blurry artifacts. Then they were split into patches and fed through our algorithm: an initial diffusion model was trained on the ``noisy'' images at f/16, and then used as a guide to learn the degradation operator on the center part of f/5.6 images.
By using the central part of f/5.6 images we should be able to fix the chromatic aberrations which appears around sharp edges but much less in central part of images.
By inspecting the data (see \cref{fig:p-lot-data}) there is a noticeable increase in blurriness between clean and noisy samples, which nevertheless is not very strong.
Unlike the previous experiments where we knew the amount of additive Gaussian noise $\sigma$ used in the forward model, now the noise distribution is completely unknown. For simplicity we again use a Gaussian noise model and treat its standard deviation as a trainable parameter of $\hatfw$. Having a good noise estimate can be very useful in guiding the final reconstruction with plug-and-play methods. 
In \cref{fig:p-lot-recon} we compare our results with the algorithm from \citet{eboli2022fast} which is a two-step approach to first estimate and remove the blur, and then remove colored fringes using a specialized procedure~\cite{Eboli2023FastCA}.
For fairness we must note that the algorithm we're comparing against~\cite{eboli2022fast} works with single images, and doesn't need retraining for different lenses, partly thanks to strong inductive priors on the the blur kernel (Gaussian with 7 parameters) and on the color aberrations. 
We also compare to commercial solution DxO PhotoLab~\citep{dxo} which exploits information about the specific lens used. We used the web version of the software and applied chromatic aberration filter followed by deblurring with strength 1.23.
More comparisons as well as training details are available in the supplementary.

\subsection{Single image super-resolution}
Up to now we have dealt with the task of deblurring, matching clean and corrupted distributions across datasets. Interestingly, there is a very related task whose properties allow us to work on single images, instead of across datasets. Super-resolution is commonly modeled as the composition of blurring with kernel $k$ and subsampling $y = (x \circledast k)\downarrow_s$, hence its close relationship to deblurring. 
In this setting our algorithm works on single low-resolution images, split into small patches. First, the noisy distribution $\dYopt$ is learned with a diffusion model on the patches. Then we learn a kernel $\hat{k}$ such that 
\begin{equation}\label{eq:sres-learn}
	p\Big((y\circledast \hat{k})\downarrow_s\Big) \approx \dYopt.
\end{equation}
Note that in \cref{eq:sres-learn} we perform a further downscaling of the already low-resolution image $y$ using the kernel $\hat{k}$ which is the target of our learning algorithm. This twice-downscaled image is compared in distribution to the once-downscaled image. Thanks to the scale-invariance of natural images the distributions (once-downscaled and twice-downscaled) will match when the kernels used for downscaling are the same, leading to the recovery of the true kernel. The downscaling step is crucial for this to work: if we omitted it (i.e. pure deblurring), we would always recover the identity kernel.
For a quantitative experiment on single image super-resolution we adopt the setting from~\citet{kligler19kernelgan} who also introduced the DIV2KRK dataset consisting of 100 images from DIV2K~\cite{agustsson17div2k} downscaled with random anisotropic Gaussian kernels. After learning $\hat{k}$ using the proposed algorithm, we tested three non-blind solvers: ZSSR~\cite{ZSSR} which requires no pretraining and USRNet~\cite{zhang2020usrnet} for which we used the \emph{tiny} pretrained model and DANv2~\cite{luo23dan} from which we took just the \emph{Restorer} module plugged with $\hat{k}$.
In \cref{tbl:sr} we compare our approach with: 
\begin{enumerate*}
    \item Two end-to-end solutions~\cite{luo23dan,luo2022deep} which jointly learn super-resolution and kernel-estimation by pre-training on DIV2K + Flickr2K~\cite{ntire2017} datasets,
    \item two non-blind super-resolution algorithms~\cite{ZSSR,zhang2020usrnet} which are natural upper-bounds for our method and 
    \item three kernel-estimation algorithms~\cite{kligler19kernelgan,yang2024dynamic,liang2021flow} which are directly comparable to the proposed method.
\end{enumerate*}
Note that both DANv2 and DCLS outperform the non-blind methods. We hypothesize this may be due to the strong prior the methods have learned during training on a dataset which is very similar to the one used in testing.
To analyze kernel estimation performance in isolation we look at kernel metrics k-PSNR and k-NCC (not available for DCLS which operates in a different kernel space) which show that the proposed algorithm performs the same as DANv2. This can be further corroborated by plugging in the learned kernels with the DANv2 super-resolution module on its own, which significantly reduces the performance gap to the end-to-end DANv2.
When compared to the other 2-step methods, the proposed algorithm emerges as the clear winner both on image and kernel metrics.

\begin{table}
   	\centering
   	\caption{x2 super-resolution on DIV2KRK dataset. Methods are grouped by the respective non-blind solver. NCC is the normalized cross-correlation metric.}
   	\label{tbl:sr}
   	\begin{tabular}{lccccc}
   		\toprule
   		Method & Y-PSNR & SSIM & LPIPS & kernel PSNR & kernel NCC \\
   		\midrule
   		\multicolumn{6}{c}{\textit{End to end (kernel estimator trained on external data)}} \\
   		DANv2~\cite{luo23dan}    & 32.12 & 0.8954 & 0.148 & 56.8 & 0.97 \\
   		DCLS~\cite{luo2022deep} & 32.31 & 0.9006 & 0.147 & -- & -- \\
   		\midrule
       	\multicolumn{6}{c}{\textit{Non-blind (true kernel available)}} \\
       	ZSSR   & 32.04 & 0.8885 & 0.196 & $\infty$ & 1 \\
       	USRNet & 32.06 & 0.8923 & 0.158 & $\infty$ & 1 \\
       	DANv2  & 31.94 & 0.8957 & 0.147 & $\infty$ & 1 \\
       	\midrule
       	\multicolumn{6}{c}{\textit{Two-step (kernel estimate from single image)}} \\
       	KernelGAN~\cite{kligler19kernelgan} & 29.65 & 0.8465 & 0.248 & 51.5 & 0.92 \\
       	Ours + ZSSR & 31.61 & 0.8837 & 0.200 & 64.2 & 0.96 \\
   		DKP~\cite{yang2024dynamic} & 23.18 & 0.6245 & 0.226 & 53.6 & 0.92 \\
   		DIP-FKP~\cite{liang2021flow} & 29.52 & 0.8530 & 0.249 & 50.4 & 0.95 \\
   		Ours + USRNet & 31.11 & 0.8838 & 0.167 & 64.2 & 0.96 \\
   		Ours + DANv2 & 31.86 & 0.8938 & 0.149 & 64.2 & 0.96 \\
   		\bottomrule
   	\end{tabular}
\end{table}
	\section{Conclusions}
	Solving blind inverse problems from unpaired data is highly relevant for practical real-world image restoration tasks. In this work, we proposed an algorithm which learns the degradation operator directly from data, making minimal assumptions about the operator itself. Through extensive experiments on uniform and non-uniform deblurring, we have shown that it is possible to reduce the performance gap between blind single-image methods and non-blind methods by leveraging multiple images from the same degradation process. Built-in interpretability and flexibility in its design make the proposed algorithm applicable to solving diverse problems. 
	On the other hand, having multiple images belonging to the same noisy-data distribution may not be realistic for all situations, where single-image methods work best -- this single-image setting can be handled with our methods only in specific problems such as super-resolution. Furthermore, while our framework is designed to deal with unpaired data, if the clean and noisy distributions are known through highly unrelated datasets, the matching procedure will struggle to learn a degradation operator to reconcile the two. The domain shift between clean and noisy distributions hinders training as it becomes larger.
	Future directions include tackling problems from different domains such as microscopy or medical imaging, as well improving the noise-modeling to be more realistic (e.g. to handle Poisson-Gaussian noise).

\subsection*{Acknowledgements}
This work was supported by ERC grant number 101087696 (APHELEIA project). This work was granted access to the HPC resources of IDRIS under the allocation AD011015445 made by GENCI.
    
	\FloatBarrier
	{
        \bibliographystyle{ieeenat_fullname}
        \bibliography{invp}
    }
 
    {
	    \begin{appendices}
%	    	\appendix
			\section{Proof of Proposition 1.}
We restate the statement of the proposition for completeness.
\begin{tprop}{}{}
	For any set of forward model parameters $\whp$, let $p_\whp(y) = \int p_\whp(y\mid x) p(x) dx$ where $p_\whp(y\mid x) = \cN(y\mid A_\whp x, \sigma^2 I)$. Let $\wopt$ be a specific set of parameters which we consider to be the optimal set.
	Then, assuming the data covariance $\Sigma = \bE_x[xx^\top]$ is invertible, there exists an orthogonal matrix $P$ such that
	\begin{equation}
		p_{\whp}(y) = p_{\wopt}(y) \implies \fw = \Sigma^{-1/2}P\Sigma^{1/2}\fwopt
	\end{equation}
%	Then for every function $f$ of the form $f(x) = x^\top C x$ for any matrix $C$, there exists an orthonormal matrix $P$ such that
%	\begin{equation}
%		\bE_{y\sim p_\whp, y'\sim p_{\wopt}} [ f(y) - f(y') ] = 0 \implies P\fw - \fwopt = 0.
%	\end{equation}
	That is, if the probability distributions $p_\whp$ and $p_{\wopt}$ are equal, it is possible to identify $\wopt$ up to rotations $P$.
\end{tprop}
\begin{proof}
	Let $f$ be a quadratic function of the form $f(x) = x^{\top}Cx$. Consider the form of the conditional, and perform a change of variables from $y$ to $Ax + \epsilon$ with $\epsilon\sim \cN(0, \sigma^2 I)$ independent of $x$ to write:
	\begin{align}
		0 &= \bE_{y\sim p_\whp}[f(y)] - \bE_{y\sim p_{\wopt}}[f(y)] \nonumber \\
		&= \int f(y) p_\whp(y\mid x) p(x) dx dy - \int f(y) p_{\wopt}(y \mid x) p(x) dx dy \nonumber \\
		&= \int f(\fw x + \epsilon) p(\epsilon) p(x) dx d\epsilon - \int f(\fwopt x + \epsilon) p(\epsilon) p(x) dx d\epsilon \nonumber \\
		&= \bE_{x\sim p(x), \epsilon} [ f(\fw x + \epsilon) - f(\fwopt x + \epsilon)]. \nonumber
	\end{align}
	Now
	\begin{align}
		\bE_{x\sim p(x), \epsilon} [ f(\fw x + \epsilon) - f(\fwopt x + \epsilon)] &= \bE_{x\sim p(x), \epsilon} [ (\fw x + \epsilon)^\top C (\fw x + \epsilon) - (\fwopt x + \epsilon)^\top C (\fwopt x + \epsilon)] \nonumber \\
		&= \bE_{x, \epsilon} [ x^\top \fw^\top C \fw x - x^\top \fwopt^\top C \fwopt x + 2 x^\top \fw^\top C \epsilon - 2 x^\top \fwopt^\top C \epsilon] \nonumber \\
		&= \bE_{x} [  x^\top \fw^\top C \fw x - x^\top \fwopt^\top C \fwopt x ] \nonumber \\
		&= \bE_{x} [ \trc(x^\top \fw^\top C \fw x) - \trc(x^\top \fwopt^\top C \fwopt x) ] \nonumber \\
		&= \bE_{x} [\trc( \fw xx^\top \fw^\top C) - \trc(\fwopt xx^\top \fwopt^\top C ) ] \nonumber \\
		&= \langle (\fw \Sigma \fw^{\top} - \fwopt \Sigma \fwopt^{\top}), C \rangle_F = 0 \nonumber \\
		&\implies \fw \Sigma \fw^{\top} - \fwopt \Sigma \fwopt^{\top} = 0 \label{eq:app-eq1}
	\end{align}
	where $\Sigma = \bE_x[xx^\top]$ is the data covariance. \Cref{eq:app-eq1} holds since $C$ can be any matrix.
	Now since $\Sigma$ is invertible, we can take $\tilde{\fw} =  \fw \Sigma^{1/2}$ and $\tilde{\fwopt} =  \fwopt \Sigma^{1/2}$ and we have that $\tilde{\fw}\tilde{\fw}^{\top} = \tilde{\fwopt} \tilde{\fwopt}^\top$.
    Now looking at the SVD of $\tilde{\fw} = U_{\whp} S_{\whp} V_{\whp}^\top$ and $\tilde{\fwopt} = U_{\wopt} S_{\wopt} V_{\wopt}^\top$ it holds that
	\begin{equation}
		U_{\whp} S_{\whp}^2 U_{\whp}^\top = U_{\wopt} S_{\wopt}^2 U_{\wopt}^\top
	\end{equation}
	which implies that both $U_{\whp} = U_{\wopt}$ and $S_{\whp} = S_{\wopt}$. 
	Take $P$ the orthonormal change of basis from $V_{\whp}$ to $V_{\wopt}$, then
	\begin{equation}
		\tilde{\fw} = \sum_i \sigma_{\whp}^{(i)} u_{\whp}^{(i)} v_{\whp}^{{(i)}^*}P = \sum_i \sigma_{\wopt}^{(i)} u_{\wopt}^{(i)} v_{\whp}^{{(i)}^*}P = \sum_i \sigma_{\wopt}^{(i)} u_{\wopt}^{(i)} v_{\wopt}^{{(i)}^*} = \tilde{\fwopt}
	\end{equation}
	and going back to the original operators, $\fw = \fwopt\Sigma^{1/2}P\Sigma^{-1/2}$ with $P$ an orthogonal matrix.
\end{proof}

\section{Experiment Details}

For all experiments we started from the diffusion model implementation EDM2~\cite{karras2024edm2}, and adapted it to train flow-matching models. The EMA for model weights based on a power-function instead of an exponential and is parameterized in an unconventional way through the relative standard deviation (see \citet{karras2024edm2} for a precised description).

\subsection{FFHQ experiment}

We used a very small CFM model with 4M parameters to train on the first 1000 samples from the FFHQ dataset, downscaled to 256x256, degraded using a fixed motion-blur kernel generated with the code from~\citet{motionblur} with intensity 0.5 and Gaussian noise with standard deviation 0.02. We trained the model from scratch for 4000 epochs, at the end of which it could generate samples like those in \cref{fig:ffhq-deg}. 
We then proceeded to the second step of our algorithm by learning a 28x28 blur kernel from 100 different clean samples from FFHQ. We enforced the kernel to sum to 1, and used a sparsity promoting regularizer (simply average of the absolute values of the kernel) with weight 1. We trained the second step for 5000 epochs with a low learning rate ($10^{-5}$ for the blur kernel and $4\times10^{-5}$ for the auxiliary CFM model). Finally we used DiffPIR~\cite{zhu23diffpir} in the implementation of DeepInv (\url{https://github.com/deepinv/deepinv}) for reconstructing the full FFHQ test set (1000 images).
The baselines were all run from their respective repositories. The only difficulty encountered was with KernelDiff~\cite{yash24kerneldiff} which cannot handle noisy measurements. We thus ran KernelDiff without noise.
\Cref{fig:ffhq,fig:ffhq-ks1} compare the image and kernel reconstructions of the different methods for both motion-blur and isotropic Gaussian kernels. Note that the same regularization and other hyperparameters were used to train our method in both experiments.

\begin{figure*}
	\centering
	\subcaptionbox{Measurement}[.16\linewidth]{\includegraphics[width=\linewidth]{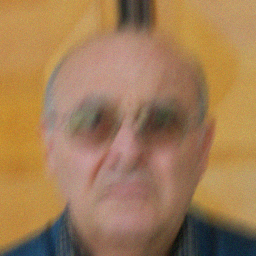}}
	\subcaptionbox{FastEM-$\Pi$GDM \cite{laroche2024fast}}[.16\linewidth]{\includegraphics[width=\linewidth]{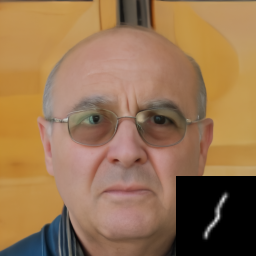}}
	\subcaptionbox{KernelDiff~\cite{yash24kerneldiff}}[.16\linewidth]{\includegraphics[width=\linewidth]{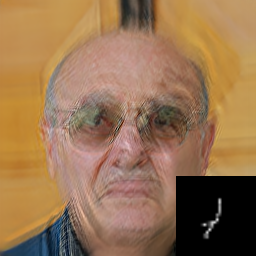}}
	\subcaptionbox{BlindDPS~\cite{chung2023parallel}}[.16\linewidth]{\includegraphics[width=\linewidth]{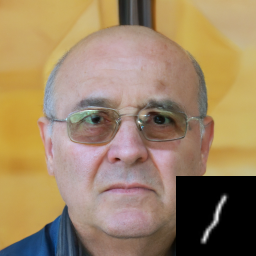}}
	\subcaptionbox{Ours}[.16\linewidth]{\includegraphics[width=\linewidth]{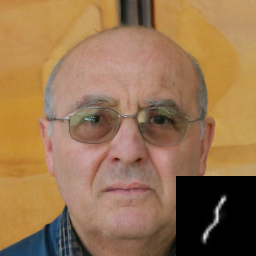}}
	\subcaptionbox{Ground truth}[.16\linewidth]{\includegraphics[width=\linewidth]{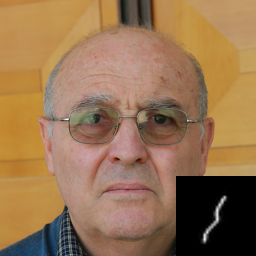}}
	\caption{Sample reconstructions on FFHQ (motion-blur kernel). All methods single-image apart from ours.}
	\label{fig:ffhq}
\end{figure*}

\begin{figure*}
	\centering
	\subcaptionbox{Measurement}[.16\linewidth]{\includegraphics[width=\linewidth]{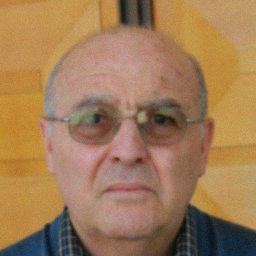}}
	\subcaptionbox{BlindDPS~\cite{chung2023parallel}}[.16\linewidth]{\includegraphics[width=\linewidth]{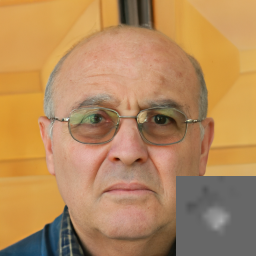}}
	\subcaptionbox{Ours}[.16\linewidth]{\includegraphics[width=\linewidth]{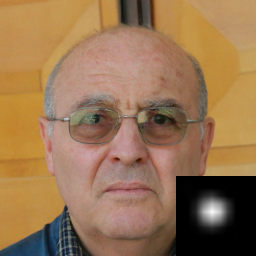}}
	\subcaptionbox{Ground truth}[.16\linewidth]{\includegraphics[width=\linewidth]{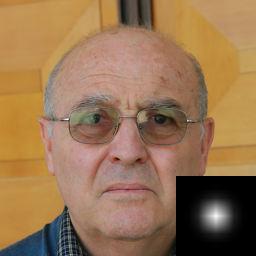}}
	\caption{Sample reconstructions on FFHQ (gaussian-blur kernel). All methods single-image apart from ours.}
	\label{fig:ffhq-ks1}
\end{figure*}

\subsubsection{FFHQ with varying noise levels}

The experiments described above and in the main text all used the same random Gaussian noise with standard deviation of 0.02. To determine whether our method is robust to different levels of noise we ran the same experiment with noise-levels 0, 0.04 and 0.08. To compare the models fairly, we then ran DiffPIR using the learned kernels (learned with different noise levels) on data which has been corrupted with the original noise level of 0.02. The training hyper-parameters for all three stages were kept fixed from the previous section.
The results of this experiment show a small decrease in accuracy as the noise-level increases, in \cref{fig:noise-varying} support a small deterioration in both kernel estimates and consequently in the reconstruction quality, while remaining fairly robust. Note that the noise is applied to 0-1 normalized images such that a noise level of 0.08 corresponds to 8\%, which is very clearly noticeable in the corrupted images.

\begin{figure}
	\begin{minipage}{0.5\linewidth}
		\includegraphics[width=0.8\linewidth]{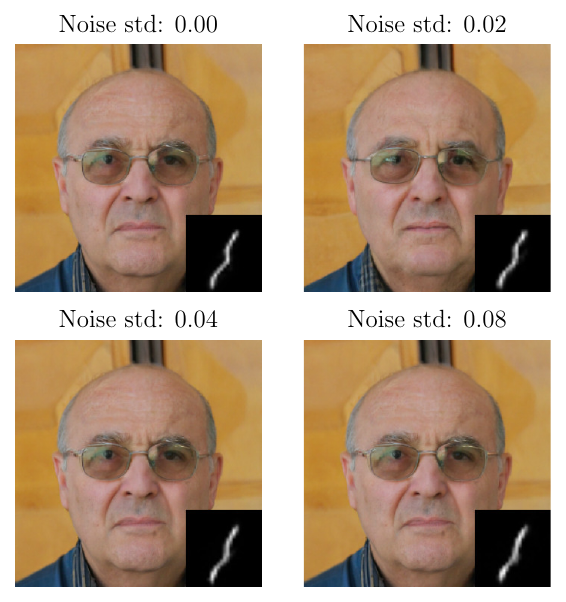}
	\end{minipage}
	\begin{minipage}{0.5\linewidth}
		\begin{tabular}{llll}
			\toprule
			$\sigma_{\mathrm{train}}$ & PSNR & LPIPS & kernel NCC \\
			\midrule
			0 & 28.93 & 0.068 & 0.98 \\
			0.02 & 28.76 & 0.069 & 0.94 \\
			0.04 & 28.84 & 0.070 & 0.96 \\
			0.08 & 28.01 & 0.079 & 0.91 \\
			\bottomrule
		\end{tabular}
	\end{minipage}
	\label{fig:noise-varying}
	\caption{Qualitative and quantitative results for motion-deblurring with increasing amounts of noise. The figure on the left shows the learned kernel and a sample reconstruction with the different noise levels. On the right, the table provides reconstruction metrics and kernel-accuracy metrics for the FFHQ test-set (1000 random images). Note that kernel NCC is the normalized cross-correlation of the learned kernel with the ground-truth kernel.}
\end{figure}

\subsection{Space-varying blur on DPDD}

Here we used a larger CFM model with 52M parameters, pretrained on the DIV2K clean training set. This then helps speed up the first step of our algorithm which consists in fine-tuning the model on the same dataset, degraded using the per-pixel blur kernels, as linearly interpolated from the 8x8 PSFs shown in \cref{fig:ddpd_kernel_err} and Gaussian noise with standard deviation 0.01. The model was trained on patches of size 128x128 from the degraded DIV2K dataset with learning rate $10^{-4}$ and exponential moving average over the weights. 
For the second step, we once again directly optimized the parameters of an 8x8 grid of kernels (the same as the ground-truth). A learning rate scheduler is used for the auxiliary flow matching model (square root scheduling with warmup). A batch size of 20 was used, and we found the batch size to significantly affect convergence speed (i.e. convergence depended on how many times each of the two models was updated, so higher batch sizes did not speed up much). For this step, we use a regularization composed of three different terms, the first one is a center term which avoid the kernels to have centering offset, the second one is a sparsity one which is basically a l1 regularization which limit the number of non zero pixels in the kernels and the last one is a Gaussian prior which  help with the overall shape of kernels. Each term has the same weight of 0.003. 
For the third step we used the DPDD dataset and perform the forward model inversion in two ways:
\begin{itemize}
    \item ESRGAN, used by generating a paired dataset on the ``train\_c'' split of DPDD. We used the implementation at \url{https://github.com/XPixelGroup/BasicSR}, changed the upscaling factor to 1 since we were not interested in super-resolution and reduced the number of iterations to 60000 since convergence was fast.
    \item DPIR, again using the implementation from DeepInv, we only need to process the DPDD test-set.
\end{itemize}

To train the Deflow~\cite{wolf21deflow} we use the official repository which is available on github at \url{https://github.com/volflow/DeFlow}.
Deflow requires a bit of preprocessing to create the training datasets. Indeed, we use both DIV2K and DPDD to train this model. So we use DPDD as a clean dataset with the low quality dataset being the DPDD training set downsampled by 4 using BICUBIC interpolation. And we use DIV2K for the noisy dataset applying a per pixel kernel interpolated from the 8x8 grid on which we add a gaussian noise of standard deviation 0.01. We also downsample the noisy DIV2K to have noisy low quality dataset.
Concerning the parameters of the training we only change the batch size in order to make it fit our GPUs and we change the scale to 4 because we only found a pretrained model for this downsampling scale.

In \cref{fig:dpdd-full}, we show the full reconstructed images used for \cref{fig:ddpd_comparison}.
The first 2 images are methods which use the real degradation which are our baselines. The 3 following images come from the other methods we compare ours to and the last image is our method.
\begin{figure}
    \centering
    \subcaptionbox{DPIR true kernels}[0.3\textwidth]{\includegraphics[width=\linewidth]{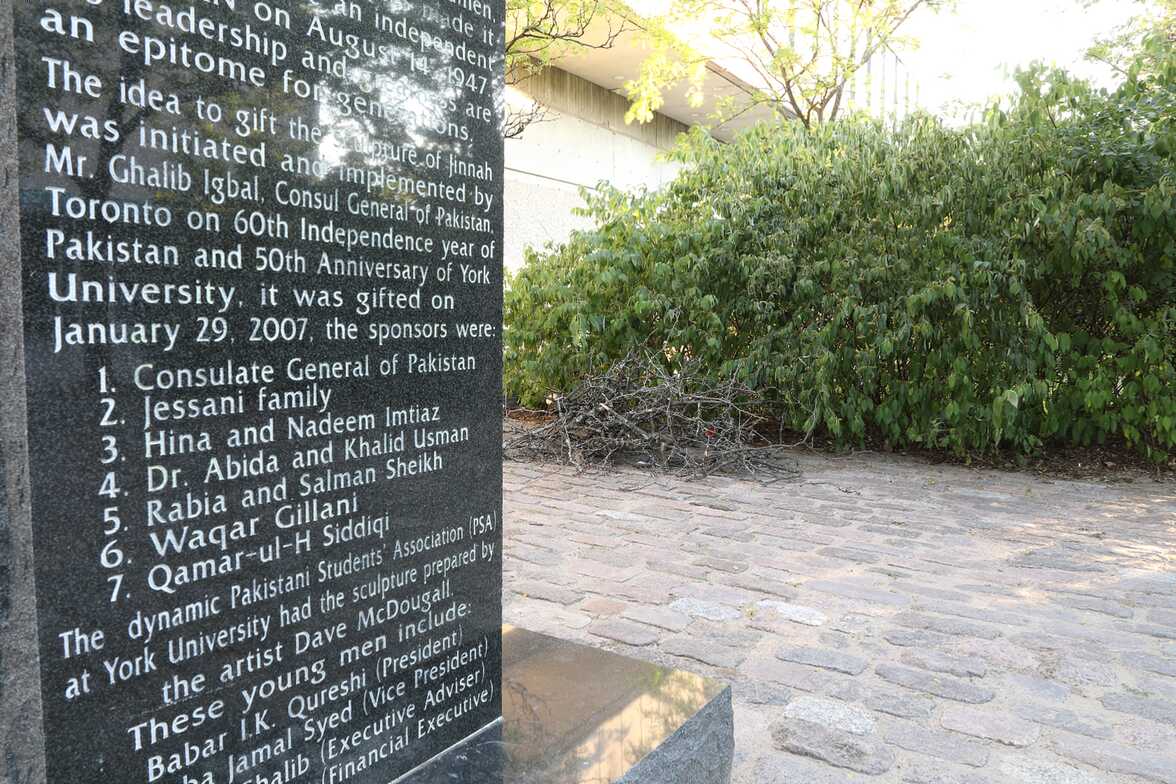}}
    \subcaptionbox{ESRGAN true kernels}[0.3\textwidth]{\includegraphics[width=\linewidth]{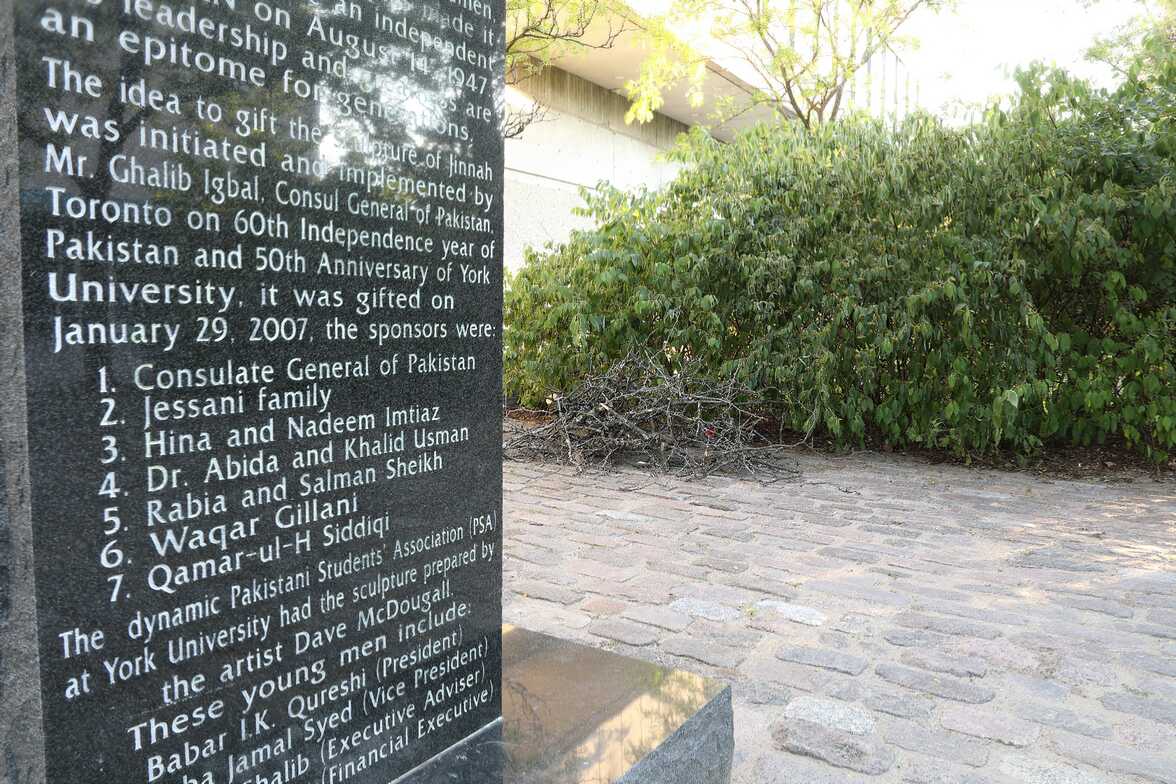}}
    \subcaptionbox{INIKNet}[0.3\textwidth]{\includegraphics[width=\linewidth]{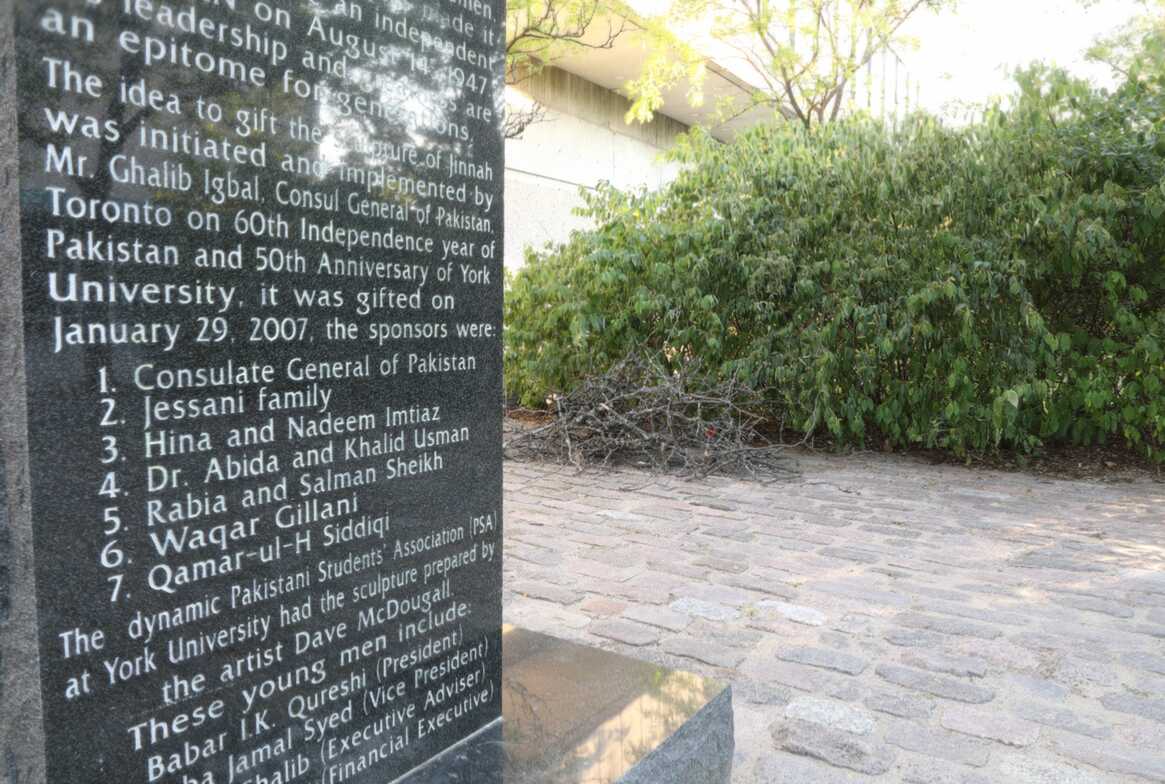}}
    \subcaptionbox{Restormer}[0.3\textwidth]{\includegraphics[width=\linewidth]{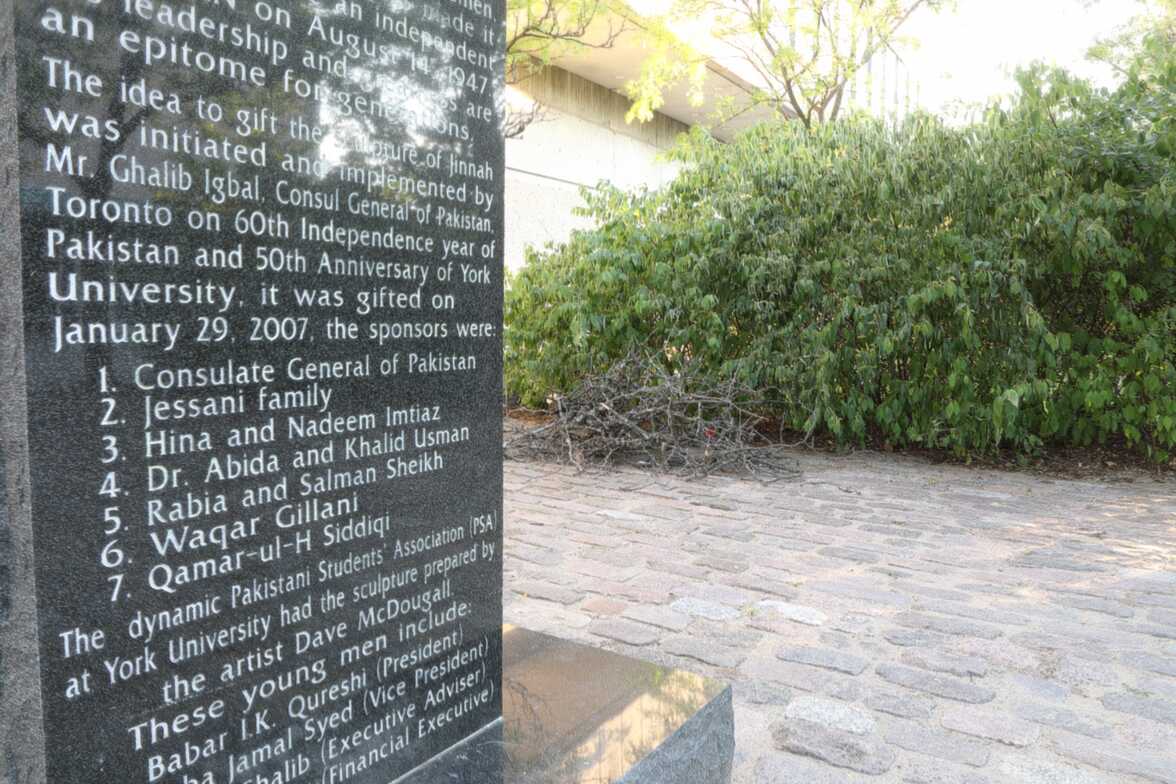}}
    \subcaptionbox{Deflow}[0.3\textwidth]{\includegraphics[width=\linewidth]{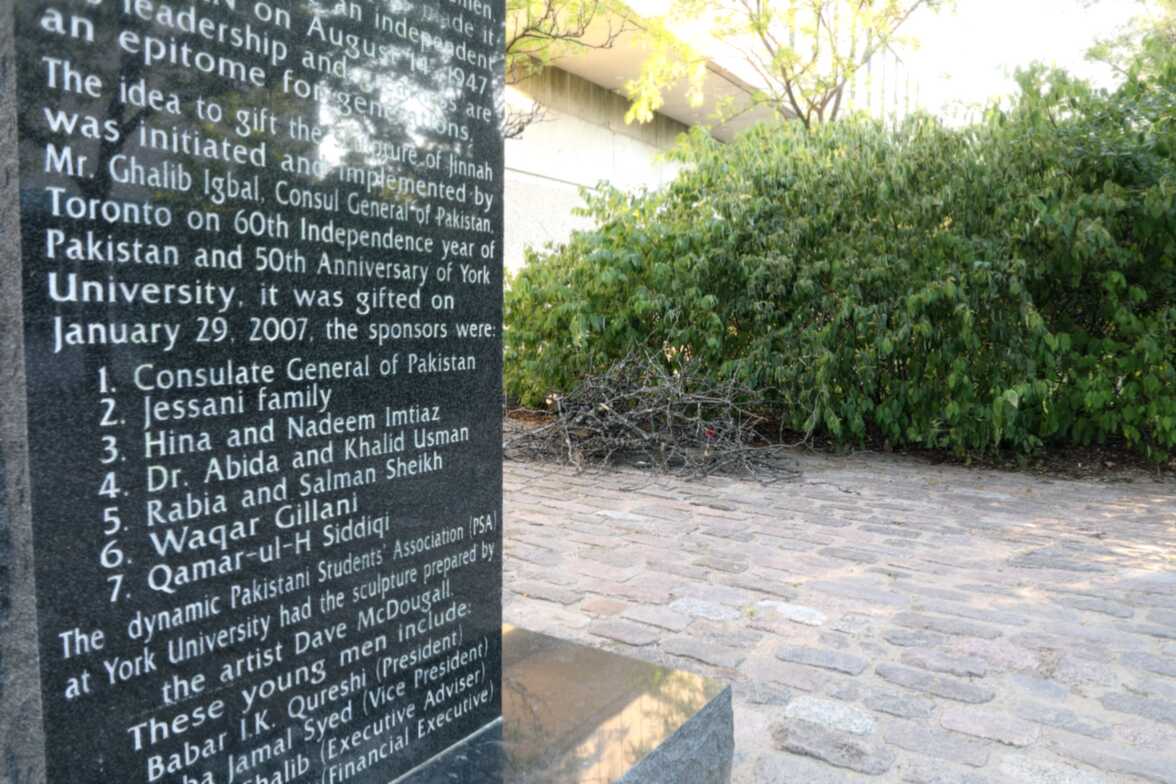}}
    \subcaptionbox{Ours + DPIR}[0.3\textwidth]{\includegraphics[width=\linewidth]{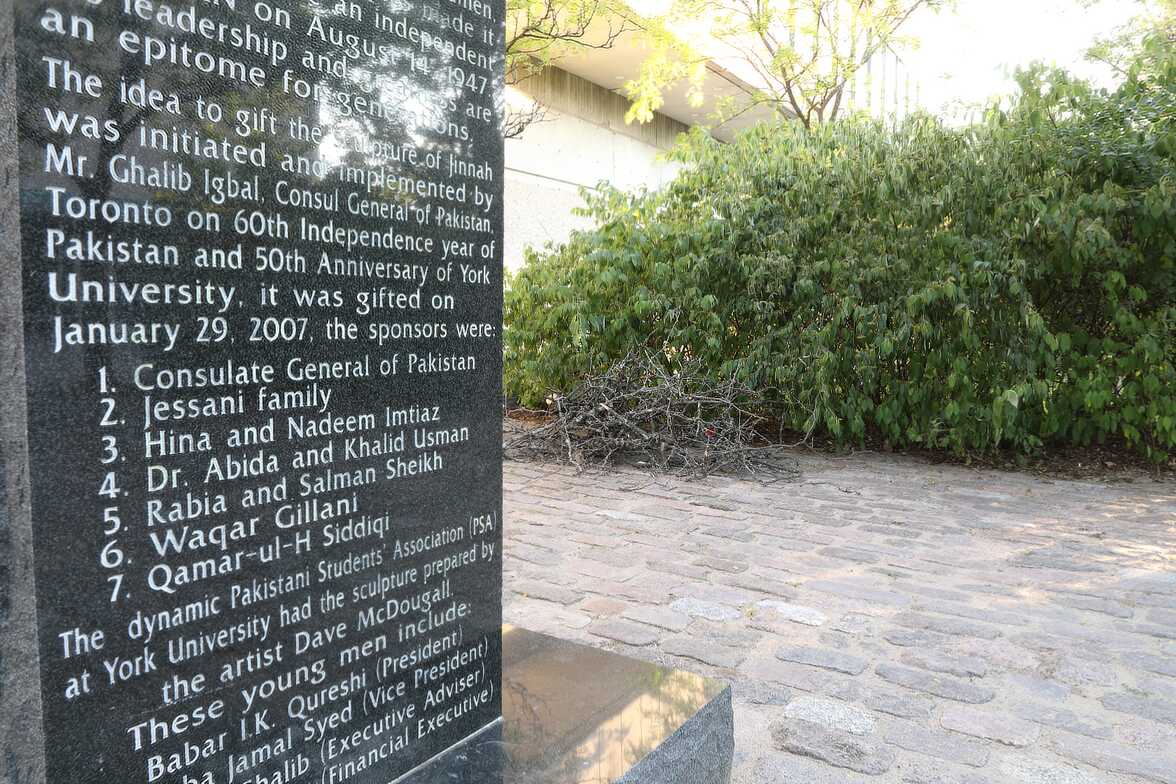}}
    \caption{Full restored DPDD image for each method analyzed}
    \label{fig:dpdd-full}
\end{figure}

\subsection{Parking-lot experiment}

\paragraph{Dataset}
We used a Panasonic DC-GX9 camera in aperture priority mode with a Leica DG Summilux 25mm f/1.4 II lens to take 22 distinct photographs in a parking lot at apertures f/16 and f/5.6. We used a tripod for stabilisation but the two images are still not perfectly aligned (for this reason we do not provide any quantitative metric on experiments using this dataset).
We process the raw images using dcraw~\cite{dcraw} to fix white balance, image highlights, demosaicking, and a small amount of noise reduction. We then converted images to png format. A low-resolution sample of the images in the dataset is shown in \cref{fig:plot_data}. The high-resolution counterparts have 5200x3904 pixels.

\begin{figure}
    \centering
    \includegraphics[width=0.3\linewidth]{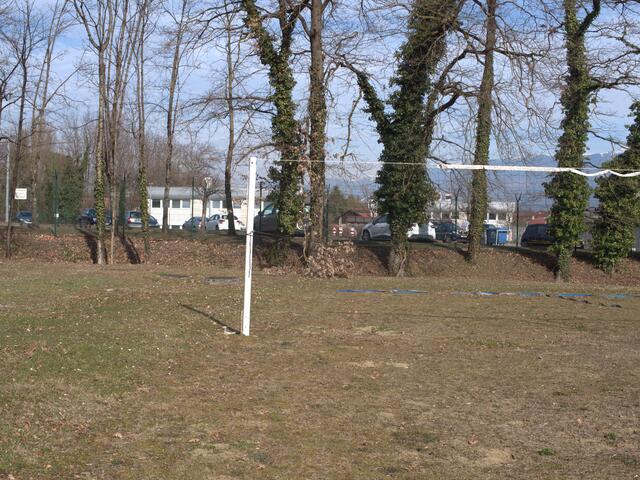}
    \includegraphics[width=0.3\linewidth]{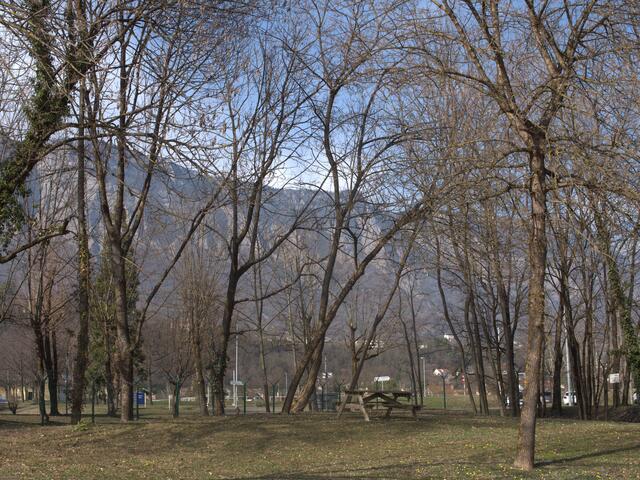}
    \includegraphics[width=0.3\linewidth]{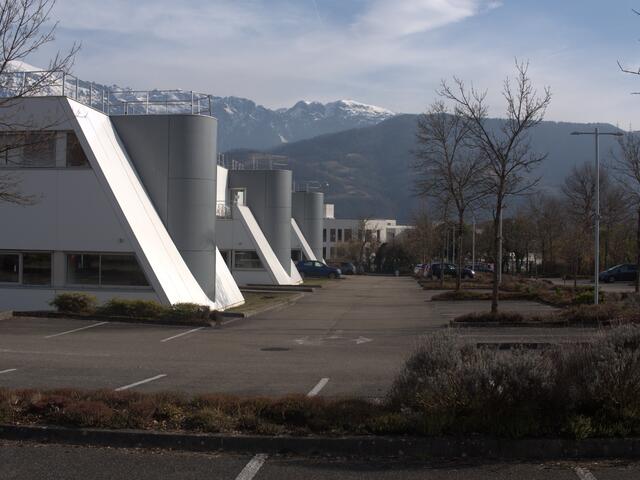}
    \includegraphics[width=0.3\linewidth]{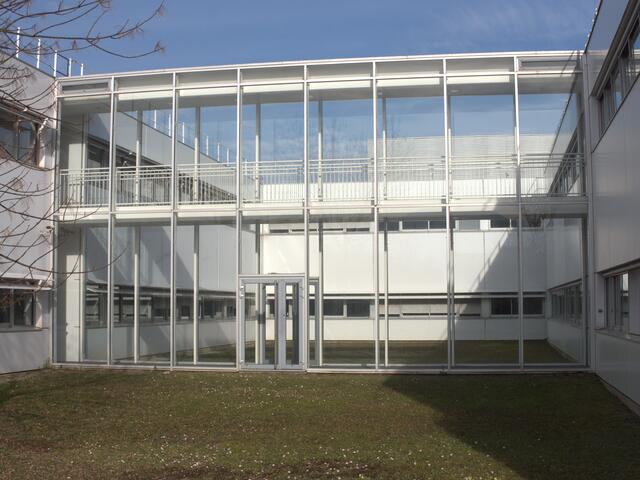}
    \includegraphics[width=0.3\linewidth]{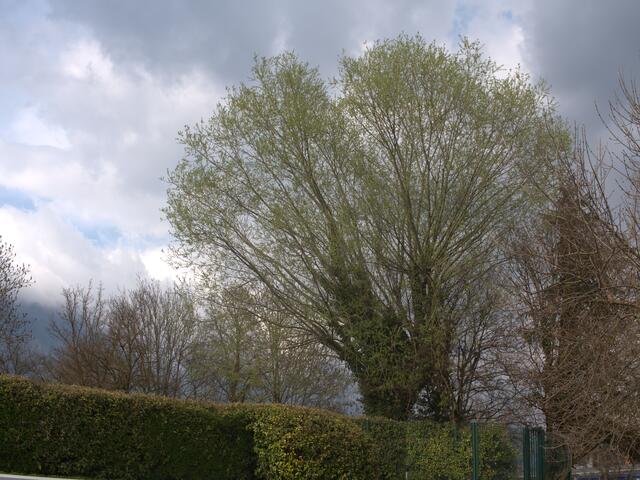}
    \includegraphics[width=0.3\linewidth]{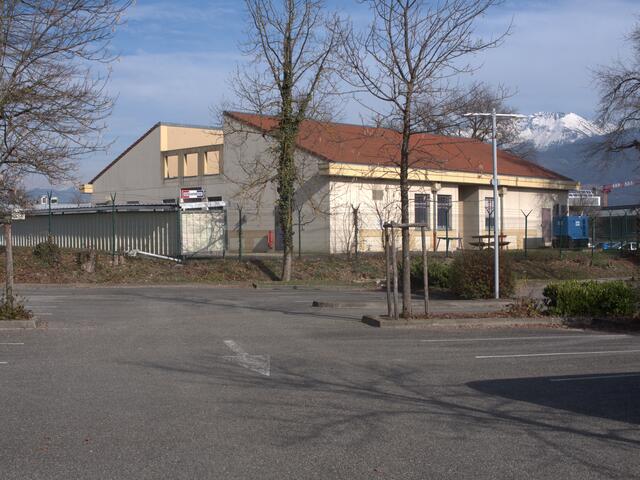}
    \caption{Samples from our parking-lot dataset. All 6 images are at f-stop 5.6}
    \label{fig:plot_data}
\end{figure}

\paragraph{Training}
We started from a scratch CFM model as the size of the one for DDPD experiment, we train it on 64-sized patches from the 18 parking lot images at f-stop 16.
For the distribution-matching step we used the the central 768x768 pixels of f/5.6 images as the clean distribution and learn a 8x8 grid of 13x13 RGB kernels. 
We compare predicted kernels with \citet{eboli2022fast} in \cref{fig:psf-ours-plot,fig:psf-eboli-plot}. 
Our PSFs are consistent across images by design, while the competing method adapts between different images. The PSFs predicted by our method more accurately recover the true blur, as seen in \cref{fig:p-lot-recon,fig:plot-additional}.

\paragraph{Hyperparameters}
The first step consists in learning a diffusion model on a single, low-resolution image. We used a flow-matching model with about 4M parameters, trained on patches of size 64x64. We used a batch-size of 1024 patches, without any added noise, learning rate of 0.01 and 0.05 as an EMA value. We trained for 16M patches.
To learn a different blur filter for different locations in the image plane, we condition the diffusion model on the original patch's location in the full image. We do this by concatenating the input image with two positioning-channels (for the x and y axes). Since the dataset is small, to avoid completely overfitting the diffusion model to the location-specific conditioning, we replace the conditioning for 20\% of the patches with a randomly chosen location.
For the second step we used a smaller batch size of 128, learning rate of 0.00001 for the auxiliary diffusion model and 0.00004 for the kernel network (described in the next paragraph). We added sparsity regularization (l1) with strength 0.01 and gaussian prior regularization with strength 0.1. The second step was trained for 6M patches.
Here we wished to learn a location-conditioned degradation which, when applied to any part of the central portion of a clean image, it would replicate the degradations of a specific location of the corrupted images. To achieve this, we conditioned the step-2 model (whose inputs are patches from the central portion of clean images) on random patch locations (spanning the full image size).
Like for DPDD we parameterized the degradations by a 8x8 grid of kernels, which unlike DPDD are different for the three RGB channels (to represent chromatic aberrations accurately). Final filters for reconstruction (which is patch-wise) are linearly interpolated from this 8x8 grid. The learned PSF grid is shown in \cref{fig:psf-ours-plot}.
We additionally added a trainable parameter to learn the noise standard deviation of the images, which was assumed to be fixed throughout the image. The noise estimate was low (around 0.003) because similar amounts of noise were present in the clean and corrupted images.
For the third step, we used plug-and-play method DPIR~\cite{zhang2021plug} which uses a deep natural image prior. We found the results were very similar to classical Wiener deconvolution, apart for some extra noise removal with DPIR. Additional qualitative reconstructions are shown in \cref{fig:plot-additional}.

\begin{figure}
    \centering
    \includegraphics[width=0.7\linewidth]{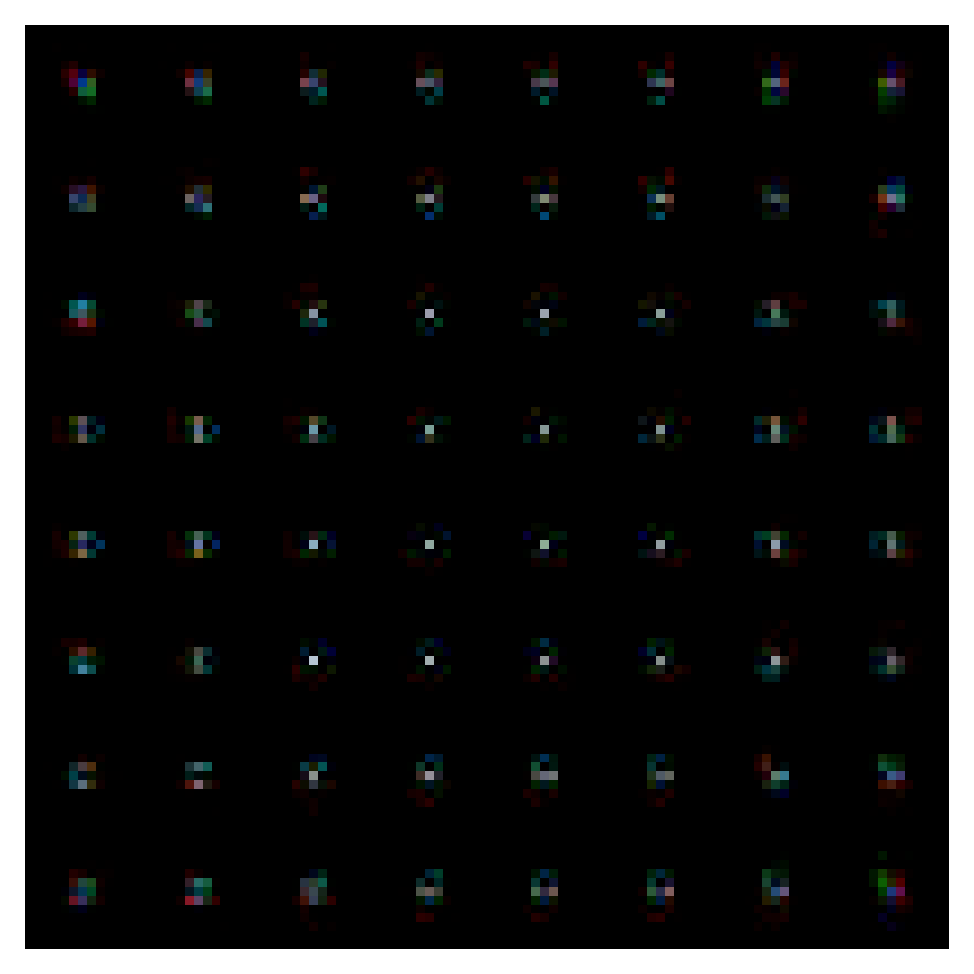}
    \caption{Lens PSF predicted by our model brightened x2. Note how the chromatic aberration (different position of red and blue channels) varies smoothly between the top, bottom and left, right parts of the lens.}
    \label{fig:psf-ours-plot}
\end{figure}
\begin{figure}
    \centering
    \includegraphics[width=0.4\linewidth]{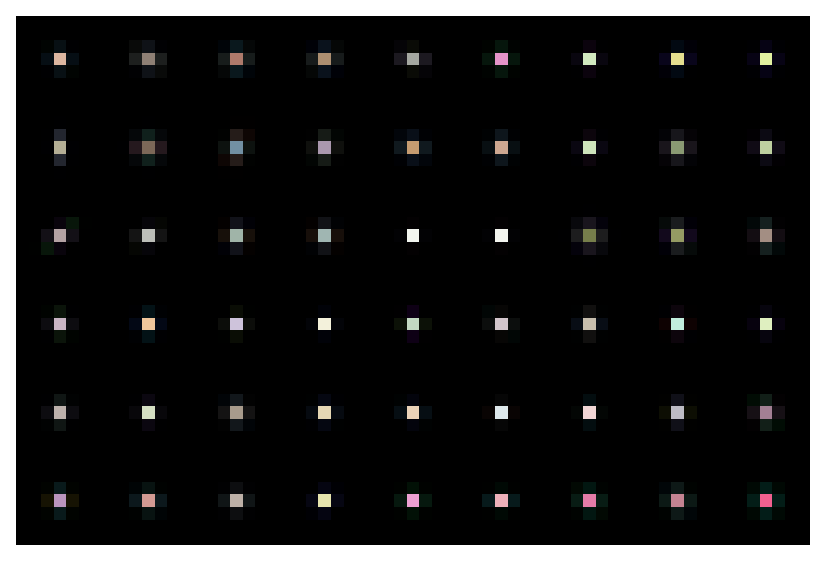}
    \includegraphics[width=0.4\linewidth]{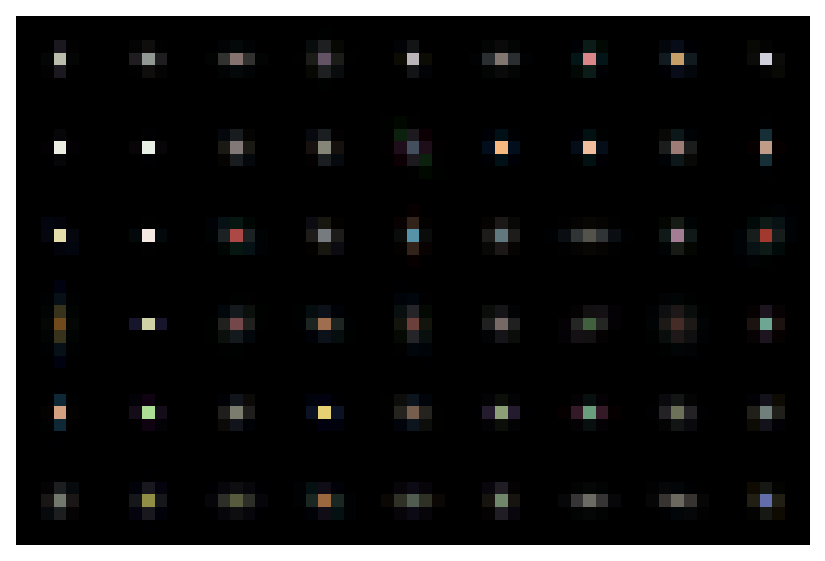}
    \caption{Lens PSF by \citet{Eboli2023FastCA} for two different images. Note that this method corrects for color aberration in a separate successive step. The PSFs here are not brightened, and hence much smaller than those in \cref{fig:psf-ours-plot}.}
    \label{fig:psf-eboli-plot}
\end{figure}

\begin{figure}
	\centering
	\includegraphics[width=\textwidth]{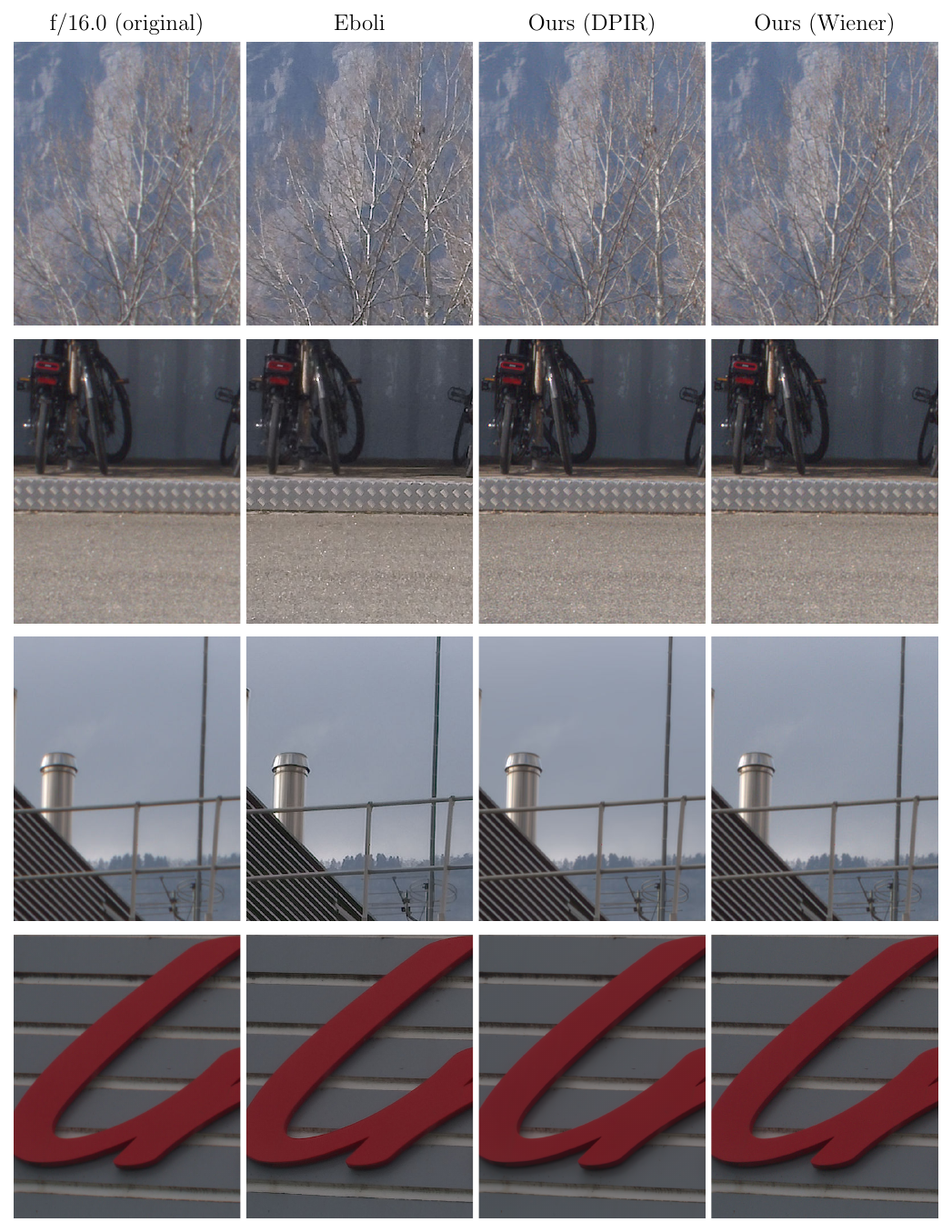}
	\caption{Additional reconstructions on parking lot data (best viewed zoomed-in).}
	\label{fig:plot-additional}
\end{figure}

\subsection{Super Resolution}

\paragraph{Hyperparameters}
The first step consists in learning a diffusion model on a single, low-resolution image. We used a small flow-matching model with about 4M parameters, trained on patches of size 32x32. We used a batch-size of 512 patches, without any added noise, learning rate of 0.01 and a small EMA value. We trained for 1M images (or approximately 2000 steps).
For the second step we used a smaller batch size of 64, learning rate of 0.00001 for the auxiliary diffusion model and 0.00008 for the kernel network (described in the next paragraph). We added center-regularization with strength 0.01, sparsity regularization (l1) with strength 0.1 and sum-to-one regularization with strength 0.1. Note that the kernel network was not forced to output normalized kernels (unlike all previous experiments), so the sum-to-one regularization was important.
The second step was trained for 512k images (or approximately 8000 steps).

\paragraph{Kernel parameterization}
Similarly to KernelGAN~\cite{kligler19kernelgan} and DCLS~\cite{luo2022deep} we use a linear convolutional network to model the blur kernel. This network can be used directly as forward model $\fw$, and can be collapsed into an explicit kernel by convolving a dirac function (i.e. an image with a single non-zero pixel in the center). We could confirm the findings of \citet{kligler19kernelgan} that this parameterization is easier to train than an equivalent CNN with a single layer. We used 4 layers with kernel-sizes 7, 5, 5 and 1, (for an overall receptive field of 15) no bias and 64 channels.

\paragraph{Additional results}
In \cref{fig:sr-comp} we show zoom-ins on four different images (35, 36, 37 and 39) from the DIV2KRK dataset as reconstructed with different methods, as well as the blur kernels predicted by the respective methods. We can see that while DANv2~\cite{luo23dan} has better reconstruction, the predicted kernels do not show significant differences to the ones predicted with our method. The better reconstruction is due to the relatively lower accuracy of ZSSR compared to the integrated super-resolver in DANv2. FKP has decent reconstruction because it always predicts blur kernels which are close to being isotropic Gaussians with a small standard deviation, while DKP predicts kernels with a higher variance which are not very close to the true ones.

In \cref{fig:sr-psnr-diff} we plot the distribution of the difference in PSNR between blind SR methods and their non-blind counterpart. Note that this can be greater than zero when due to random stochasticity of the non-blind method, the estimated kernel results in a better reconstruction than the true one.

\begin{figure}
	\centering
	\includegraphics[width=\textwidth]{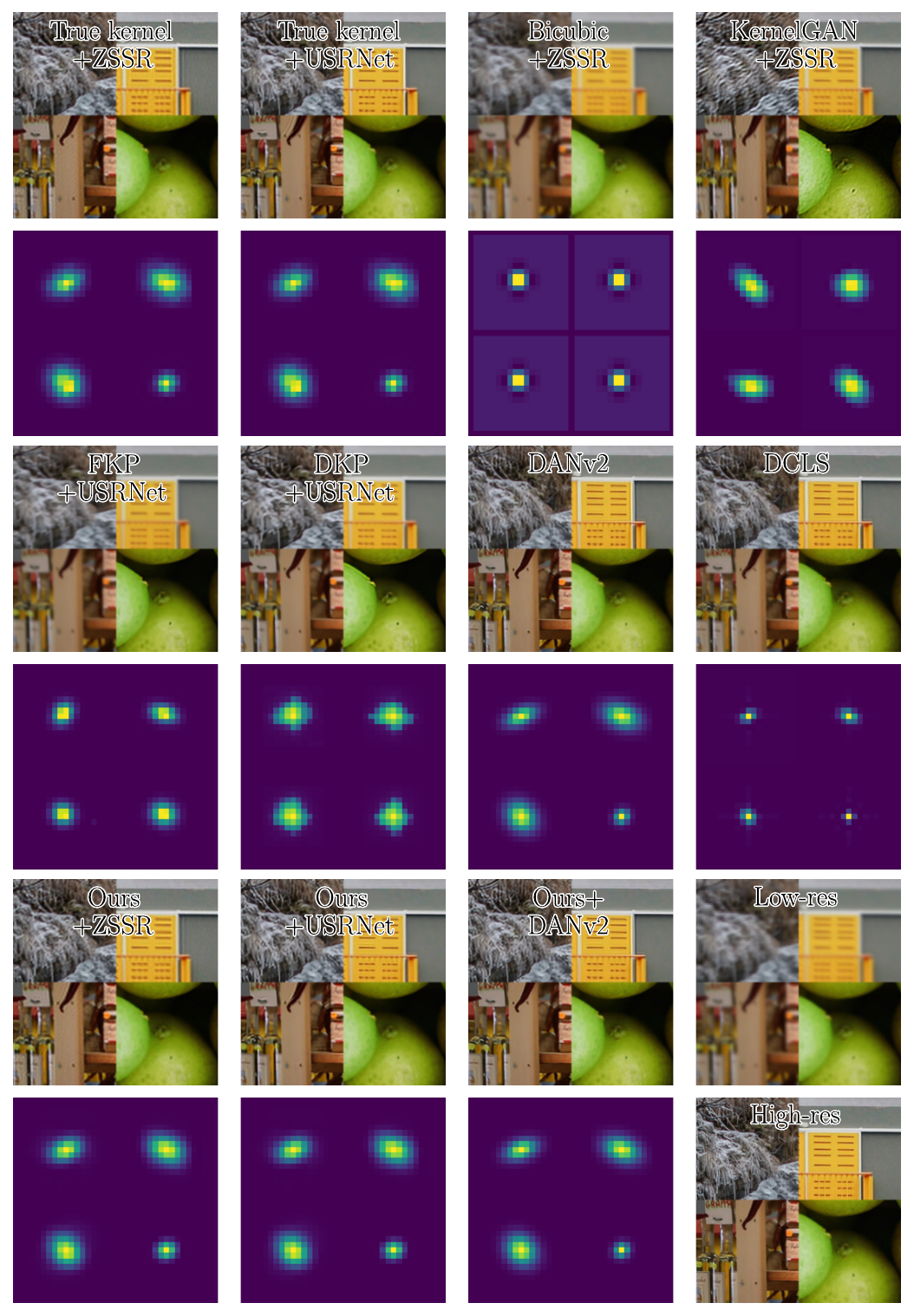}
	\caption{Super-resolution: qualitative results. Note that DCLS kernels are not comparable as they live in a different space.}
	\label{fig:sr-comp}
\end{figure}

\begin{figure}
	\centering
	\includegraphics[width=0.6\textwidth]{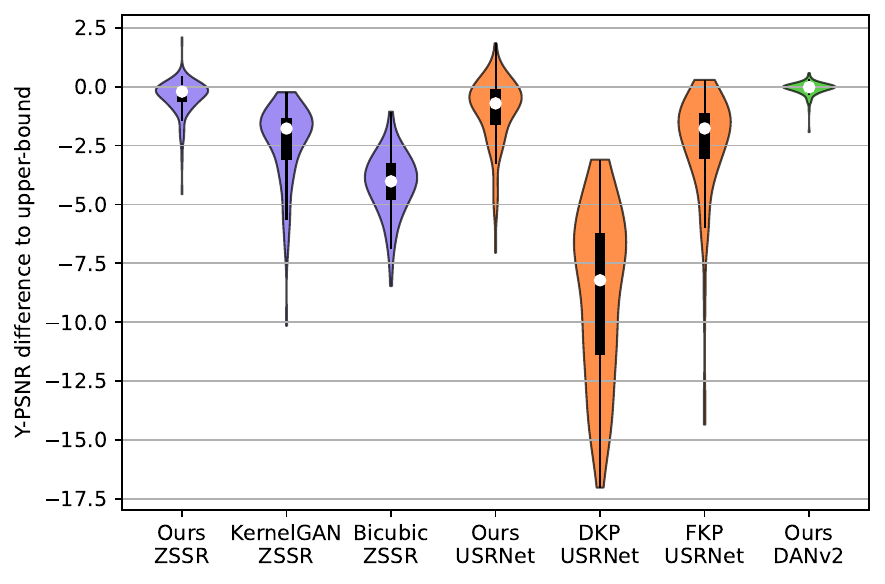}
	\caption{Super-resolution: Performance gap of two-step methods compared to their natural upper-bound. The metric considered is the Y-PSNR (PSNR on the luminance only).}
	\label{fig:sr-psnr-diff}
\end{figure}

\paragraph{Metrics}
We specify here the parameters used to calculate metrics in \cref{tbl:sr}. Firstly, we found USRNet to produce very strong border artifacts which -- if included -- would have completely skewed the results. For this reason we crop 60 pixels off each edge for every image before computing metrics. For the PSNR, in order to maintain consistency with results reported in the literature, we used the Y-PSNR (i.e. computed on just the luminance channel, after converting RGB images to YCbCr). For the LPIPS metric we used the ''alex-net`` network.
The PSNR on kernels was computed after padding the kernels with zeros up to size 25x25 and after shifting them so they were appropriately centered. The NCC metric is the 2d normalized cross-correlation, which can better account for small centering errors in the kernels.

	    \end{appendices}
	    
	}
    
\end{document}